\documentclass[10pt,twocolumn,letterpaper]{article}

\usepackage[pagenumbers]{cvpr} 

\usepackage{amsmath}  
\usepackage{graphicx}
\usepackage{bm}
\usepackage{multirow}
\usepackage{makecell}
\usepackage{cuted}
\usepackage{caption}

\definecolor{cvprblue}{rgb}{0.21,0.49,0.74}
\usepackage[pagebackref,breaklinks,colorlinks,allcolors=cvprblue]{hyperref}

\def\confName{CVPR}
\def\confYear{2027}

\title{MixReact: Physics-Based Reactive Humanoid Control from Multimodal Cues via a Discrete–Continuous Mixture of Transformers}

\author{
Bin Li$^*$ \textsuperscript{1} \quad 
Xing Zhu$^*$ \textsuperscript{1} \quad 
Jingya Wang \textsuperscript{\textdagger} \textsuperscript{1,2} \\
\textsuperscript{1} \textit{ShanghaiTech University} \quad 
\textsuperscript{2} \textit{InstAdapt}\\
}

\begin{document}
\maketitle
\renewcommand{\thefootnote}{\fnsymbol{footnote}}
\footnotetext[1]{Equal contribution.}
\footnotetext[2]{Corresponding author.}
\begin{strip}
\centering
\includegraphics[width=\textwidth]{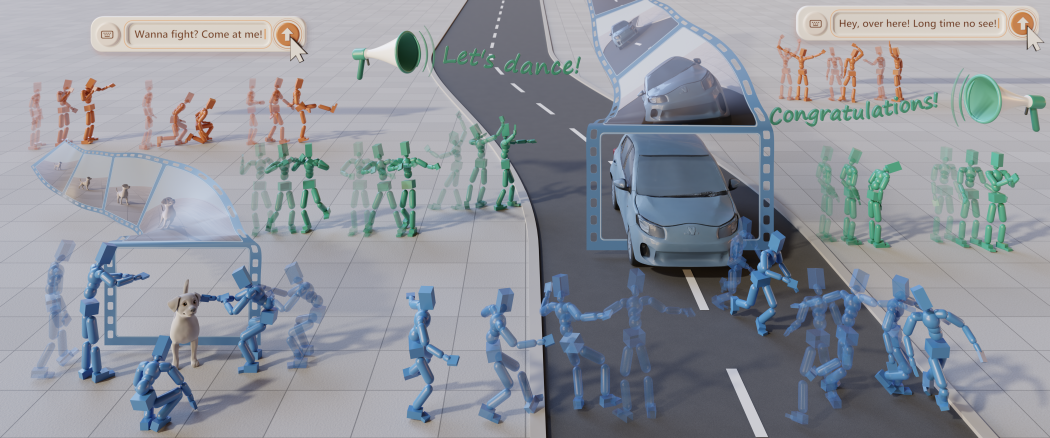}
\captionsetup{hypcap=false}
\captionof{figure}{\textbf{MixReact} executes diverse humanoid reactions to everyday multimodal cues in physics-based simulation. Examples include combat and greeting behaviors driven by text (orange), bowing, and applause elicited by audio (green), and interactions with an approaching puppy or avoidance of an oncoming vehicle prompted by video (blue). Multiple behaviors per cue illustrate one-to-many reaction control.}
\captionsetup{hypcap=true}
\label{fig:teaser}
\end{strip}
\begin{abstract}
Humanoid agents in everyday environments are expected to react to diverse cues as humans do, with semantically appropriate and physically executable behaviors. Existing cue-driven reaction generation methods support only a narrow range of cue modalities and are limited to purely kinematic generation, leaving the resulting reactions susceptible to physical artifacts. To this end, we present \textbf{MixReact}, the first unified framework for physics-based reactive humanoid control from text, audio, and video cues, integrating a Discrete--Continuous Mixture of Transformers (DisCo-MoT) for state prediction with a History-aware Action Transformer (HAT) for physical execution. DisCo-MoT couples a discrete masked-modeling branch with a continuous autoregressive diffusion branch via one-way mixture attention within a MoT architecture. Guidance from discrete context mitigates collapse into averaged behaviors in one-to-many reaction generation, while direct prediction in continuous state space bypasses the decoding bottleneck of discrete representations. HAT then maps the predicted next state and observed execution history to executable actions, enabling robust closed-loop control in physics-based simulation. Experiments across all three modalities demonstrate consistent improvements over the baselines in fidelity, diversity, and physical consistency, while ablations validate the complementary contributions of the components. Our code will be released to facilitate future research. Project page: \small
    \url{https://binlee26.github.io/MixReact_Page/}
\end{abstract}    
\section{Introduction}
\label{sec:intro}
Humans rarely act in isolation. We step back when a dog lunges, raise an arm against an incoming ball, and nod or wave when someone greets us. Endowing humanoid agents with human-like responsiveness to everyday cues is a core challenge in computer vision, with applications spanning animation, gaming, VR/AR, and embodied AI.

Most reaction generation methods~\cite{ji2025towards,chopin2023interaction,ghosh2024remos,yi2025interactive,xu2024regennet,tan2025think,cen2025ready,liu2024physreaction,men2022gan,goel2022interaction,jiang2025arflow} condition on a partner's body trajectory. This confines reactions to human--human interaction and presumes clean 3D motion of the counterpart. Recent work broadens this scope: HERO~\cite{yu2025hero} introduces video-driven reaction generation, extending interactions to animals and environmental events, while NOUS~\cite{zhou2026nousvideodriven3dhuman} improves the correspondence between visual observations and generated reactions. In parallel, ReactMotion~\cite{luo2026reactmotion} supports text- and audio-driven reactions through ReactMotionNet, a large-scale dataset pairing speaker utterances with listener behaviors. Despite this progress, these works focus solely on kinematic generation without physical constraints, leaving generated motions prone to artifacts such as penetration, foot sliding, and floating that compromise realism and practical applicability. Moreover, each supports only a subset of cue modalities, and no existing framework unifies text-, audio-, and video-driven reactive control. To close both gaps, we introduce \emph{physics-based reactive humanoid control from multimodal cues} (Fig.~\ref{fig:teaser}): given a text, audio, or video cue, the goal is to control a humanoid to produce a semantically appropriate, physically plausible reaction.

The central challenge lies in the one-to-many mapping from cues to reactions: a greeting, for example, may elicit a wave, a nod, or a bow.  Most existing physics-based  methods~\cite{tevet2025closd,wu2025uniphys,li2026interagent,li2026mind,truong2024pdp,huang2025diffuse,yao2022controlvae,ren2023insactor,yuan2023physdiff,luo2024universal} model behavior in continuous spaces, a formulation that may cause distinct reaction modes to blend into an averaged response under one-to-many cue-to-reaction mappings. Discrete token-based modeling~\cite{yao2024moconvq,pinyoanuntapong2024bamm,jiang2023motiongpt,zhu2026motiongpt3,zhang2023generating,guo2022tm2t,guo2024momask} mitigates this collapse by sampling distinct reaction modes from categorical token distributions, but quantization-induced reconstruction errors compromise the fine-grained contact and balance information essential for physics-based execution. We observe that the two paradigms are complementary: which reaction to perform is naturally a discrete choice, while how to execute it requires precise detail. 

\begin{figure*}
    \centering
    \includegraphics[width=1.0\textwidth]{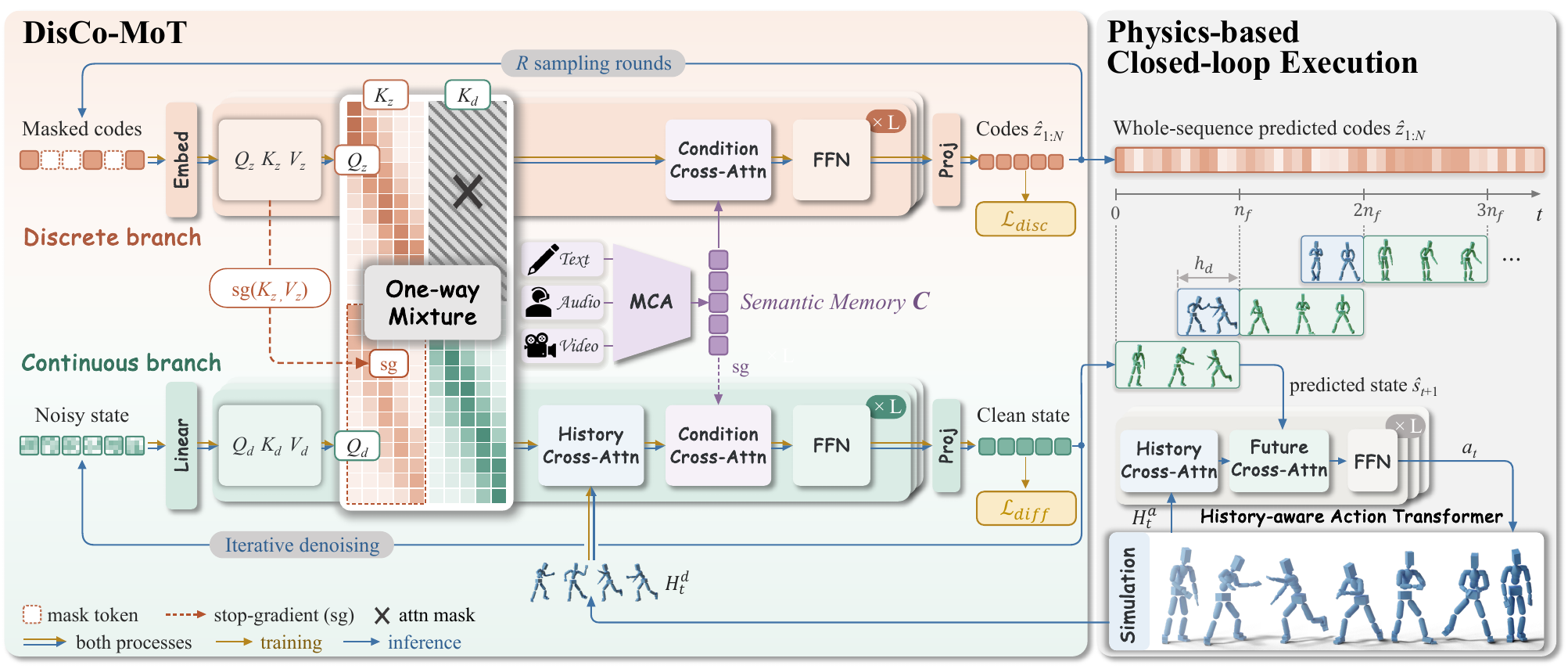}
    \caption{\textbf{Overview of MixReact.} MCA maps a text, audio, or video cue to the semantic memory $C$. Within DisCo-MoT, discrete masked modeling samples a complete reaction once, providing cached intermediate keys and values that guide history-conditioned autoregressive diffusion through one-way mixture attention. HAT converts each predicted next state and observed execution history into an executable action, while simulator feedback updates both prediction and control histories.}
    \label{fig:overview}
\end{figure*}

To this end, we propose \textbf{MixReact}, a unified physics-based reactive control framework that takes text, audio, or video cues and drives a humanoid to produce physically plausible and semantically consistent reactions. At its core is a \textbf{Dis}crete--\textbf{Co}ntinuous \textbf{M}ixture \textbf{o}f \textbf{T}ransformers (DisCo-MoT), a novel state generation architecture that couples a discrete masked-modeling branch~\cite{pinyoanuntapong2024mmm,chang2022maskgit,guo2024momask,pinyoanuntapong2025maskcontrol,li2025lamp} with a continuous autoregressive diffusion branch~\cite{tevet2025closd,xiao2025motionstreamer,zhao2025dartcontrol,song2020denoising,nichol2021improved,ho2020denoising,han2024amd,zhao2026autoregressive,chen2024taming,shi2024interactive,li2024aamdm} within a MoT~\cite{zhou2025transfusion,liang2024mixture} framework, as shown in Fig.~\ref{fig:overview}. Specifically, the discrete branch samples a complete reaction as a sequence of learned state tokens that serves as persistent global behavioral context. The continuous branch accesses this context through one-way mixture attention over key--value caches from the discrete branch’s intermediate Transformer layers, while keeping the discrete context independent of denoising features. Conditioned on this context and actual execution history, it autoregressively predicts future states in continuous space. This design mitigates mode collapse through guidance from the discrete context while directly modeling behavior in continuous space, thereby circumventing the reconstruction bottleneck associated with decoding tokenized representations~\cite{van2017neural,razavi2019generating,lee2022autoregressive}.
To enable robust physical execution, we introduce a \textbf{H}istory-aware \textbf{A}ction \textbf{T}ransformer ({HAT}) that maps the predicted next state and observed execution history to physically executable actions. This establishes a state-based interface between reaction generation and closed-loop control. Extensive experiments demonstrate that MixReact consistently outperforms the baselines in fidelity, diversity, and physical consistency. 

In summary, our contributions are as follows:
\begin{itemize}
    \item We propose \text{MixReact}, the first unified framework for physics-based reactive humanoid control from text, audio, and video cues, integrating \text{DisCo-MoT} for reaction generation with \text{HAT} for closed-loop physical execution.

    \item We introduce \text{DisCo-MoT}, a state generation architecture that couples discrete masked modeling with continuous autoregressive diffusion via one-way mixture attention. Guidance from discrete context mitigates mode collapse, while prediction in continuous state space circumvents the decoding bottleneck of discrete representations.

    \item We develop HAT, a history-aware physics-based controller, which maps the predicted state and observed execution history to actions for robust closed-loop control.

\end{itemize}

\section{Related Work}
\label{sec:related_work}

\paragraph{Physics-based humanoid control.}
DeepMimic~\cite{peng2018deepmimic} learns imitation policies from reference motions, AMP~\cite{peng2021amp} introduces adversarial motion priors, and PHC~\cite{luo2023perpetual} scales tracking with failure recovery. Reusable representations span continuous~\cite{peng2022ase,won2022physics,yao2022controlvae,tessler2023calm,dou2023c,luo2024universal} and discrete~\cite{zhu2023neural,yao2024moconvq} latent spaces for downstream control. PhysDiff~\cite{yuan2023physdiff}, InsActor~\cite{ren2023insactor}, and CLoSD~\cite{tevet2025closd} couple motion generation with physical execution. Complementary approaches explore language-directed control~\cite{juravsky2022padl}, motion inpainting~\cite{shrestha2024generating,tessler2024maskedmimic}, progressive policy distillation~\cite{juravsky2024superpadl}, and diffusion-based control~\cite{truong2024pdp,zhang2026script,wu2025uniphys,huang2025diffuse,ye2026predactor,li2026interagent,li2026mind}. However, physics-based control of one-to-many reactions to everyday multimodal cues remains underexplored.

\paragraph{Human reaction control.}
Reaction synthesis commonly conditions on a partner's 3D motion~\cite{chopin2023interaction,ghosh2024remos,xu2024regennet,tan2025think,xu2025moreact,javed2024intermask,zhang2025reactffusion,beg2026reactionmamba}. Recent work broadens conditioning to RGB or egocentric video~\cite{yu2025hero,zhou2026nousvideodriven3dhuman,zhang2025egoreact}, speaker utterances~\cite{luo2026reactmotion}, and spatial audio~\cite{xu2026mospa,xu2026mosat,ni2026sonic}. Online approaches employ causal generation~\cite{yi2025interactive,cen2025ready}, reduce error accumulation through training on generated histories~\cite{geng2026armflow}, or incorporate others' actions and scene geometry~\cite{ye2026remogen}. PhysReaction~\cite{liu2024physreaction} and Human-X~\cite{ji2025towards} support physical execution through dynamics-guided imitation and actor-aware tracking, respectively, but primarily require a partner's 3D motion. Spatial-audio methods also demonstrate downstream physics-based tracking~\cite{xu2026mospa,ni2026sonic}. MixReact unifies text-, audio-, and video-driven reactions with closed-loop physical execution, without requiring a partner's 3D motion, which is often difficult to acquire.

\paragraph{Mixture of Transformers.}
Transfusion~\cite{zhou2025transfusion} combines next-token prediction with denoising diffusion~\cite{ho2020denoising,song2020denoising,nichol2021improved} in a shared Transformer. MoT~\cite{liang2024mixture} assigns modality-specific parameters while retaining global attention, with related architectures in LMFusion~\cite{shi2026lmfusion} and BAGEL~\cite{deng2025emerging}. Vision-language-action models constrain information flow through asymmetric attention~\cite{black2026pi0visionlanguageactionflowmodel,intelligence2025pi05visionlanguageactionmodelopenworld} or specialist-to-generalist querying~\cite{yu2026twinbrainvla}. LLaMo~\cite{li2026llamo} and MotionGPT3~\cite{zhu2026motiongpt3} extend language models with MoT and continuous motion representations. DisCoRD~\cite{cho2025discord} maps discrete tokens to continuous motion through rectified flow. DisCo-MoT instead couples discrete masked modeling~\cite{pinyoanuntapong2024mmm,chang2022maskgit,guo2024momask,pinyoanuntapong2025maskcontrol,li2025lamp,pinyoanuntapong2024bamm,yuan2024mogents,li2026remomask,hwang2026snapmogen} with continuous autoregressive diffusion~\cite{tevet2025closd,xiao2025motionstreamer,zhao2025dartcontrol,han2024amd,zhao2026autoregressive,chen2024taming,shi2024interactive,li2024aamdm,li2024autoregressive,chen2024diffusion,zhang2025primal} through one-way mixture attention, allowing a complete discrete reaction plan to guide continuous state prediction.
\section{Method}
\label{sec:method}

\begin{table*}
\caption{Comparison of text- and audio-driven physics-based humanoid reaction  control on the ReactMotionNet~\cite{luo2026reactmotion} test set, where $\pm$ indicates a 95\% confidence interval and $\rightarrow / \uparrow / \downarrow$ means closer / larger / smaller is better. \textbf{Bold} and \underline{underline} indicates the best and the second-best results, respectively. Phys-GT refers to the physics-based trajectories that are tracked from the MoCap dataset.}
\centering
\resizebox{\textwidth}{!}{%
\begin{tabular}{clcccccc}
\toprule
 &  & FID $\downarrow$ & Diversity $\rightarrow$ & MModality $\uparrow$ & Floating $[mm]$ $\downarrow$ & Skating $[mm]$ $\downarrow$ & Jerk $[mm/frame^3]$ $\downarrow$\\
\midrule
 & Phys-GT          & $0.000^{\pm 0.000}$ & $1.359^{\pm 0.046}$ & - &$13.559^{\pm 0.000}$ & $1.064^{\pm 0.000}$ &  $1.976^{\pm 0.000}$\\
\cmidrule(lr){2-8}
\multirow{3}{*}{Text}
 & ReactMotion~\cite{luo2026reactmotion}      & $1.478^{\pm 0.002}$ & $1.176^{\pm 0.019}$ & $1.035^{\pm 0.021}$ & $29.213^{\pm 0.269}$ & $2.523^{\pm 0.015}$ & $\mathbf{2.279}^{\pm 0.010}$\\
 & ReactMotion++~\cite{luo2026reactmotion,luo2023perpetual}    & $\underline{1.379}^{\pm 0.005}$ & $\underline{1.301}^{\pm 0.011}$ & $\underline{1.208}^{\pm 0.035}$ & $19.035^{\pm 0.659}$ & $1.560^{\pm 0.080}$ & $4.372^{\pm 0.354}$ \\
 & CLoSD~\cite{tevet2025closd}    & ${1.543}^{\pm 0.010}$ & $1.091^{\pm 0.017}$ & $0.443^{\pm 0.034}$ & $\underline{16.493}^{\pm 0.119}$ & $\underline{1.341}^{\pm 0.029}$ & ${3.045}^{\pm 0.022}$ \\
 & MixReact (Ours)  & $\mathbf{1.328}^{\pm 0.005}$ & $\mathbf{1.356}^{\pm 0.003}$ & $\mathbf{1.320}^{\pm 0.003}$ & $\mathbf{13.827}^{\pm 0.086}$ & $\mathbf{0.917}^{\pm 0.028}$ & $\underline{2.626}^{\pm 0.036}$ \\
\midrule
\multirow{3}{*}{Audio}
 & ReactMotion~\cite{luo2026reactmotion}       & $1.515^{\pm 0.011}$ & $1.175^{\pm 0.039}$ & $1.165^{\pm 0.094}$ & $28.193^{\pm 0.257}$ & $2.352^{\pm 0.095}$ & $\mathbf{2.280}^{\pm 0.024}$ \\
 & ReactMotion++~\cite{luo2026reactmotion,luo2023perpetual}    & $\underline{1.386}^{\pm 0.010}$ & $\underline{1.292}^{\pm 0.009}$ & $\underline{1.290}^{\pm 0.052}$ & $23.534^{\pm 0.791}$ & $2.369^{\pm 0.135}$ & $7.662^{\pm 0.623}$ \\
 & CLoSD~\cite{tevet2025closd}   & ${1.543}^{\pm 0.011}$ & $1.090^{\pm 0.005}$ & $0.451^{\pm 0.049}$ & $\underline{16.552}^{\pm 0.070}$ & $\underline{1.329}^{\pm 0.022}$ & ${3.080}^{\pm 0.042}$ \\
 & MixReact (Ours)  & $\mathbf{1.326}^{\pm 0.006}$ & $\mathbf{1.361}^{\pm 0.003}$ & $\mathbf{1.324}^{\pm 0.003}$ & $\mathbf{13.920}^{\pm 0.061}$ & $\mathbf{0.987}^{\pm 0.028}$ & $\underline{2.645}^{\pm 0.034}$  \\
\bottomrule
\end{tabular}%
}
\label{table:compare_baseline_reactmotion}
\end{table*}
\begin{table*}
\caption{Comparison of video-driven physics-based humanoid reaction  control on the ViMo~\cite{yu2025hero} test set.}
\centering
\resizebox{\textwidth}{!}{%
\begin{tabular}{clcccccc}
\toprule
 &  & FID $\downarrow$ & Diversity $\rightarrow$ & MModality $\uparrow$ & Floating $[mm]$ $\downarrow$ & Skating $[mm]$ $\downarrow$ & Jerk $[mm/frame^3]$ $\downarrow$\\
\midrule
 & Phys-GT          & $0.000^{\pm 0.000}$ & $1.402^{\pm 0.006}$ & - &$22.378^{\pm 0.000}$ & $3.851^{\pm 0.000}$ &  $18.623^{\pm 0.000}$ \\
\cmidrule(lr){2-8}
\multirow{4}{*}{Video}
 &HERO~\cite{yu2025hero} & $1.036^{\pm 0.003}$ & $1.314^{\pm 0.006}$ & $0.786^{\pm 0.010}$ & $38.036^{\pm 1.000}$ & $9.642^{\pm 0.126}$ & $\mathbf{4.477}^{\pm 0.050}$  \\
&NOUS~\cite{zhou2026nousvideodriven3dhuman}   & $1.026^{\pm 0.003}$ & $1.321^{\pm 0.007}$ & $0.773^{\pm 0.005}$ & $39.004^{\pm 0.584}$ & $10.959^{\pm 0.078}$ & $\underline{4.661}^{\pm 0.047}$\\
&HERO++~\cite{yu2025hero,luo2023perpetual}  & $0.862^{\pm 0.004}$ & $\underline{1.367}^{\pm 0.005}$ & $\underline{1.005}^{\pm 0.007}$ & $28.173^{\pm 0.211}$ & $\underline{4.600}^{\pm 0.098}$ & $10.242^{\pm 0.229}$ \\
&NOUS++~\cite{zhou2026nousvideodriven3dhuman,luo2023perpetual}   & $\underline{0.856}^{\pm 0.005}$ & $1.366^{\pm 0.007}$ & $1.002^{\pm 0.014}$ & $29.471^{\pm 0.144}$ & $4.683^{\pm 0.177}$ & $11.217^{\pm 0.091}$  \\
&CLoSD~\cite{tevet2025closd}  & $0.888^{\pm 0.002}$ & $1.362^{\pm 0.014}$ & $0.843^{\pm 0.029}$ & $\underline{23.669}^{\pm 0.196}$ & $5.021^{\pm 0.146}$ & ${8.869}^{\pm 0.268}$  \\
&MixReact (Ours)  & $\mathbf{0.706}^{\pm 0.003}$ & $\mathbf{1.382}^{\pm 0.003}$ & $\mathbf{1.244}^{\pm 0.003}$ & $\mathbf{20.908}^{\pm 0.185}$ & $\mathbf{3.081}^{\pm 0.062}$ & ${7.505}^{\pm 0.075}$ \\
\bottomrule
\end{tabular}%
}
\label{table:compare_baseline_vimo}
\end{table*}
Given a text, audio, or video cue $c_m$, MixReact generates a semantically appropriate reaction and executes it in physics-based simulation (Fig.~\ref{fig:overview}). The multimodal conditioning adapter (MCA) first constructs a semantic memory $C$ (Sec.~\ref{subsec:MCA}), which conditions reaction state generation in the Discrete--Continuous Mixture of Transformers (DisCo-MoT; Sec.~\ref{subsec:MoT}). Within DisCo-MoT, the discrete masked-modeling branch samples a complete reaction as a sequence of state tokens, providing persistent global behavior context for the continuous autoregressive diffusion branch. Through one-way mixture attention, this context guides future-state prediction alongside observed execution history, while remaining independent of denoising features. The History-aware Action Transformer (HAT) then combines each denoised next state with observed execution history to generate an executable action (Sec.~\ref{subsec:HAT}). Simulator feedback updates the execution history, closing the loop between reaction generation and physical control.

\subsection{Multimodal Conditioning Adapter.}\label{subsec:MCA}
MCA maps heterogeneous text, audio, and video features into a common semantic memory $C$ through modality-specific adapters and a shared projection, providing a unified conditioning representation for DisCo-MoT. For modality $m$, a frozen encoder extracts $F_m=\mathcal{E}_m(c_m)\in\mathbb{R}^{L_m\times d_m}$, where $L_m$ and $d_m$ denote the token count and feature dimension. Each token is mapped as
\begin{equation}
\begin{aligned}
E_{m,i} &= \operatorname{LN}_m(W_m F_{m,i}+b_m+p_{m,i})+e_m, \\
C_{m,i} &= W_C\operatorname{Dropout}(E_{m,i})+b_C.
\end{aligned}
\end{equation}
Here, $W_m$, $b_m$, and $\operatorname{LN}_m$ are modality-specific; $p_{m,i}$ and $e_m$ encode position and modality identity, respectively; and $(W_C,b_C)$ is shared across modalities. Projected sequences are concatenated into $C$, with a key-padding mask restricting attention to valid tokens of the selected modality for each sample.

\subsection{Discrete-Continuous Mixture of Transformers}\label{subsec:MoT}

As shown in the left panel of Fig.~\ref{fig:overview}, DisCo-MoT couples discrete modeling of complete reactions with continuous prediction of state evolution. The discrete masked-modeling branch provides persistent behavior context, while the continuous diffusion branch predicts future states from this context and execution history. The branches maintain separate projections and feed-forward networks and communicate through one-way mixture attention.

\paragraph{Discrete masked-modeling branch.}
To maintain behavioral consistency across prediction windows, the discrete branch represents a complete reaction as a sequence of state tokens. A pretrained, frozen VQ-VAE tokenizer~\cite{van2017neural}, with codebook size $K$ and temporal downsampling factor $\kappa$, maps a $T$-frame state sequence to $Z=(z_1,\ldots,z_N)$, where $N=\lceil T/\kappa\rceil$ and $z_i\in\{1,\ldots,K\}$. Following masked modeling~\cite{chang2022maskgit,guo2024momask,pinyoanuntapong2024mmm}, the discrete branch predicts masked tokens conditioned on the unmasked tokens and semantic memory $C$. At inference, it samples a complete token sequence once at the start of each rollout and retains it unchanged throughout execution. The continuous branch accesses this context through one-way mixture attention over key--value caches from the discrete branch's intermediate Transformer layers, providing persistent global guidance across successive prediction windows. Within one-way mixture attention, discrete queries are masked from attending to continuous keys and values, while the continuous branch accesses the discrete branch's key--value caches under stop-gradient, thereby preventing continuous denoising from disrupting the global behavioral structure.

\paragraph{Continuous autoregressive diffusion branch.}
The continuous branch predicts future states directly, bypassing the decoding bottleneck of discrete representations. Let $s_t\in\mathbb{R}^{D_s}$ denote a humanoid state at frame $t$. The continuous branch predicts a window of $n_f$ future states from the preceding history $H_t^{d}=[s_{t-h_d+1},\ldots,s_t]$, where $h_d$ is the history length. At diffusion step $k$, the noisy future window is $x_k\in\mathbb{R}^{n_f\times D_s}$. At each layer, state queries jointly attend to their own branch's keys and values and cached counterparts from the corresponding discrete Transformer layer through one-way mixture attention. The discrete context guides every denoising step and persists across prediction windows, promoting behavioral consistency throughout autoregressive generation. Each layer further incorporates execution history $H_t^{d}$ and semantic memory $C$ through cross-attention, followed by a feed-forward network. The resulting clean states $\hat{x}_0$ are subsequently supplied to HAT. 

\paragraph{One-way mixture attention.}
For the continuous branch, one-way mixture attention provides reaction context through intermediate key--value caches computed from the complete token sequence $Z$. At layer $\ell$, continuous queries attend jointly to these cached features and the branch's own keys and values, with stop-gradient applied to the discrete caches:
\begin{equation}
A_d^\ell =
\operatorname{softmax}\left(
\frac{
Q_d^\ell
[\operatorname{sg}(K_z^\ell),K_d^\ell]^\top
}{\sqrt{d_h}}
\right)
[\operatorname{sg}(V_z^\ell),V_d^\ell],
\end{equation}
where $Q_d^\ell$, $K_d^\ell$, and $V_d^\ell$ are the continuous queries, keys, and values; $K_z^\ell$ and $V_z^\ell$ are the cached discrete keys and values; $d_h$ is the attention head dimension; $[\cdot,\cdot]$ denotes token concatenation; and $\operatorname{sg}$ denotes stop-gradient.
For the discrete branch, an attention mask prevents discrete queries from attending to continuous keys and values:
\begin{equation}
A_z^\ell =
\operatorname{softmax}\left(
\frac{
Q_z^\ell[K_z^\ell,K_d^\ell]^\top
}{\sqrt{d_h}} + M_z
\right)
[V_z^\ell,V_d^\ell],
\end{equation}
where $Q_z^\ell$ denotes the discrete queries and the mask is defined as
\begin{equation}
(M_z)_{ij} =
\begin{cases}
0, & 1 \leq j \leq N,\\
-\infty, & j > N,
\end{cases}
\end{equation}
with the first $N$ positions corresponding to discrete tokens. The mask excludes denoising features from the discrete branch, while stop-gradient blocks diffusion gradients through its cached keys and values, thereby preventing continuous denoising from disrupting the behavioral structure.

\subsection{History-aware Action Transformer}\label{subsec:HAT}

HAT enables robust closed-loop physical execution by conditioning action generation on the predicted next state and observed execution history. At control step $t$,
\begin{equation}
a_t=\pi(H_t^{a},\hat{s}_{t+1}),
\end{equation}
where $H_t^{a}=[s_{t-h_a+1},\ldots,s_t]$ contains the latest $h_a$ simulated states and $\hat{s}_{t+1}$ is the predicted next state supplied by DisCo-MoT. A learned action query attends to these two inputs through separate cross-attention operations, and the resulting representation is projected to the action space.
We train HAT through DAgger-based online distillation~\cite{ross2011reduction} from a motion-tracking teacher policy~\cite{luo2023perpetual}. History conditioning allows HAT to account for temporal state evolution, enabling more robust closed-loop execution. Additional details are provided in the supplementary material.

\begin{figure*}
    \centering
    \includegraphics[width=1.0\textwidth]{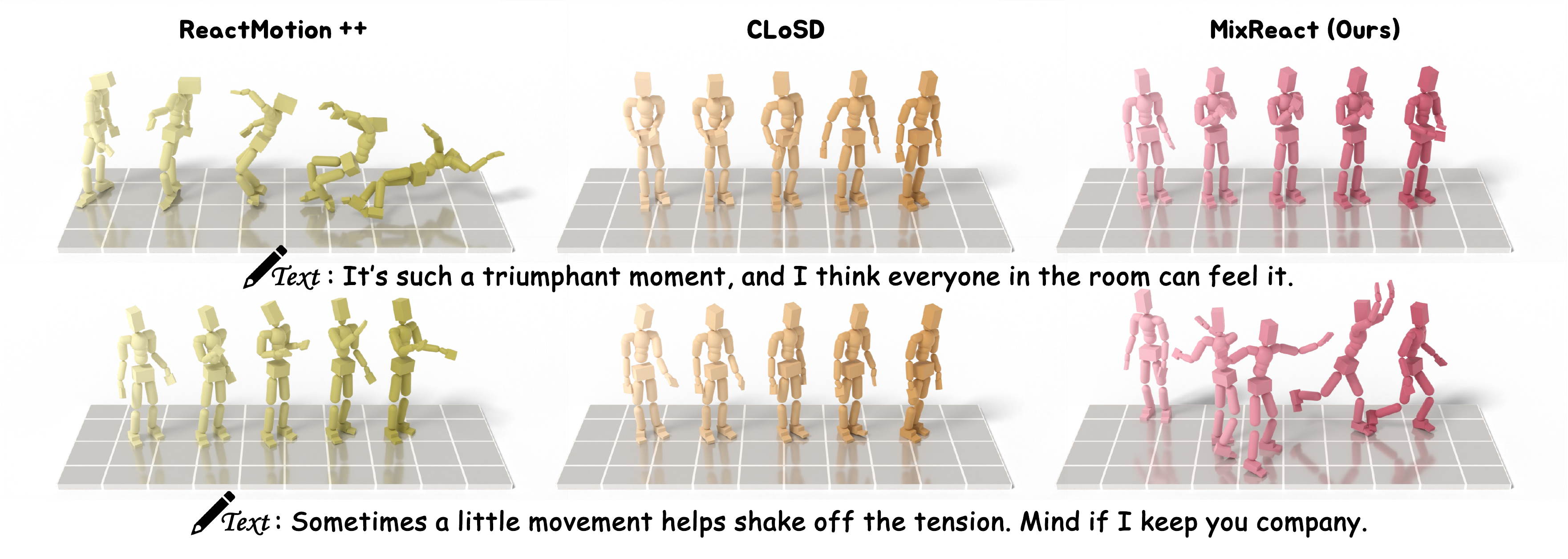}
    \caption{\textbf{Qualitative comparison of text-driven humanoid reaction control.} MixReact responds with celebratory clapping and expressive dancing. ReactMotion++~\cite{luo2026reactmotion,luo2023perpetual} loses balance in the first example and exhibits limited articulation in the second, whereas CLoSD~\cite{tevet2025closd} produces restrained responses in both. Color shading indicates temporal progression, from light (earlier) to dark (later ).}
    \label{fig:comp_text}
\end{figure*}
\begin{figure*}
    \centering
    \includegraphics[width=1.0\textwidth]{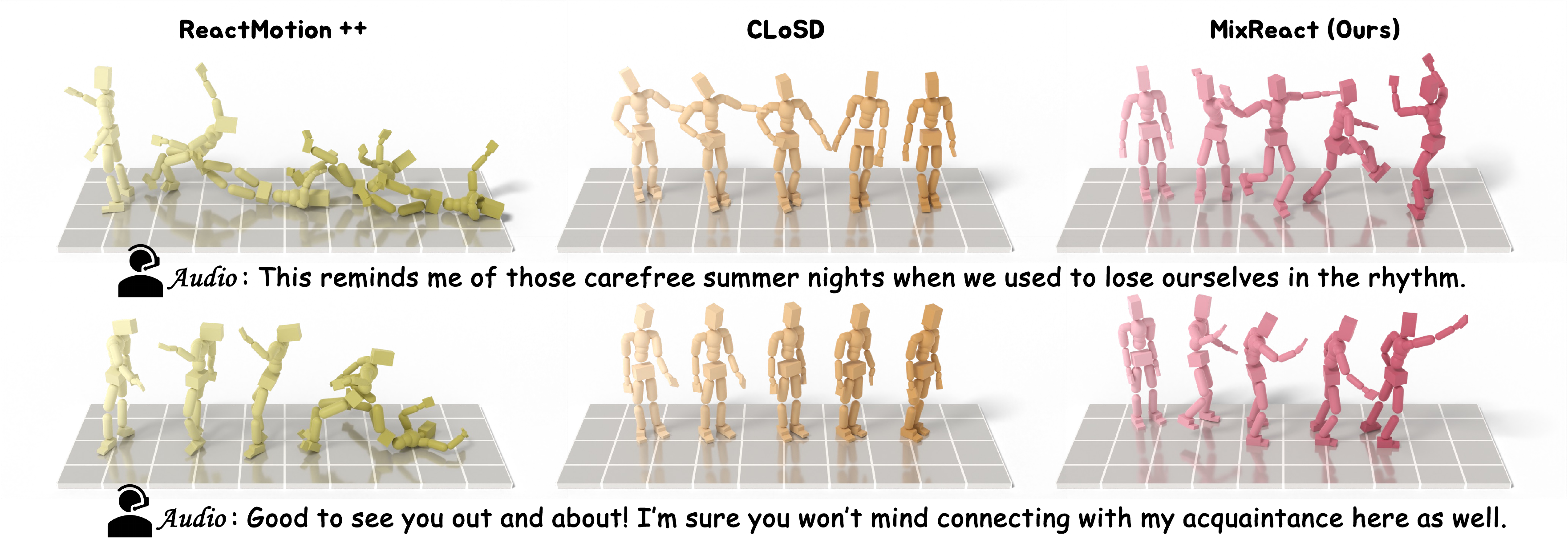}
    \caption{\textbf{Qualitative comparison of audio-driven humanoid reaction control.} MixReact produces expressive reactions while maintaining balance, whereas ReactMotion++~\cite{luo2026reactmotion,luo2023perpetual} loses balance and CLoSD~\cite{tevet2025closd} exhibits comparatively restrained responses.}
    \label{fig:comp_audio}
\end{figure*}
\subsection{Training and Inference}\label{subsec:TI}

\paragraph{Training.}
DisCo-MoT is trained with separate passes over complete reaction sequences for the discrete branch and local prediction windows for the continuous branch. Let $\theta$ denote the parameters of MCA and the discrete masked-modeling branch, and $\phi$ those of the continuous autoregressive diffusion branch.

For discrete modeling, each valid state token is independently masked with probability $p_{\mathrm{mask}}$. The branch learns cue-conditioned reaction structure through
\begin{equation}
\mathcal{L}_{\mathrm{disc}}
=
-\frac{1}{|\mathcal{M}|}
\sum_{i\in\mathcal{M}}
\log p_\theta(z_i\mid Z_{\bar{\mathcal{M}}},C),
\end{equation}
where $\mathcal{M}$ and $\bar{\mathcal{M}}$ are the sets of masked and unmasked valid token positions, respectively, and $Z_{\bar{\mathcal{M}}}$ contains the unmasked tokens.

For continuous modeling, the history $H_t^{d}$ and future window $x_0=[s_{t+1},\ldots,s_{t+n_f}]$ are extracted from ground-truth trajectories. We sample a diffusion step $k$ and perturb the future window according to a cosine noise schedule:
\begin{equation}
x_k=\sqrt{\bar{\alpha}_k}x_0+
\sqrt{1-\bar{\alpha}_k}\epsilon,
\qquad \epsilon\sim\mathcal{N}(0,I).
\end{equation}
The continuous branch attends to key--value caches from the discrete branch's intermediate Transformer layers through one-way mixture attention. Conditioned on this discrete context, semantic memory $C$, and history $H_t^{d}$, it predicts the clean future window $\hat{x}_0$ and minimizes
\begin{equation}
\mathcal{L}_{\mathrm{diff}}
=
\frac{1}{|\mathcal{V}|D_s}
\sum_{j\in\mathcal{V}}
\left\|\hat{x}_{0,j}-x_{0,j}\right\|_2^2,
\end{equation}
where $\mathcal{V}$ indexes the valid future frames within the window.

The two branches operate at complementary temporal scales: the discrete branch learns reaction semantics and global behavioral structure over complete sequences, while the continuous branch autoregressively denoises future-state windows to recover fine-grained details. Persistent discrete guidance supports behavioral consistency across windows, and conditioning on execution history promotes temporal coherence. To keep the global structure governed by full-sequence supervision, we stop diffusion gradients through both semantic memory $C$ and the discrete branch’s intermediate key--value caches, ensuring that MCA and the discrete branch are optimized exclusively by the masked-modeling objective $\mathcal{L}_{\mathrm{disc}}$. The overall objective is
\begin{equation}
\begin{aligned}
\mathcal{L}_{\mathrm{total}}
=\mathcal{L}_{\mathrm{disc}}(\theta)+\mathcal{L}_{\mathrm{diff}}\!\left(
\phi;
\operatorname{sg}(\mathcal{K}_\theta(Z,C_\theta))
\right),
\end{aligned}
\end{equation}
where $C_\theta$ denotes the semantic memory produced by MCA and $\mathcal{K}_\theta$ collects the discrete branch's keys and values across layers. Separate optimizers and data passes update $\theta$ with $\mathcal{L}_{\mathrm{disc}}$ and $\phi$ with $\mathcal{L}_{\mathrm{diff}}$, respectively.

\begin{table*}
\caption{\textbf{Ablation study of MixReact.} We evaluate the contributions of key components across all three cue modalities.}
\centering
\resizebox{\textwidth}{!}{%
\begin{tabular}{clcccccc}
\toprule
 &  & FID $\downarrow$ & Diversity $\rightarrow$ & MModality $\uparrow$ & Floating $[mm]$ $\downarrow$ & Skating $[mm]$ $\downarrow$ & Jerk $[mm/frame^3]$ $\downarrow$\\
\midrule
\multirow{4}{*}{Text}
 & MixReact (Ours)  & $\mathbf{1.328}^{\pm 0.005}$ & $\underline{1.356}^{\pm 0.003}$ & $\underline{1.320}^{\pm 0.003}$ & $\underline{13.827}^{\pm 0.086}$ & $\underline{0.917}^{\pm 0.028}$ & $\underline{2.626}^{\pm 0.036}$   \\
 & w/o Discrete branch   & $1.514^{\pm 0.011}$ & $1.178^{\pm 0.007}$ & $1.001^{\pm 0.060}$ & $\mathbf{10.808}^{\pm 0.063}$ & $\mathbf{0.133}^{\pm 0.006}$ & $\mathbf{1.025}^{\pm 0.028}$ \\
 & w/o Continuous branch & $1.423^{\pm 0.020}$ & $1.306^{\pm 0.020}$ & $1.280^{\pm 0.014}$ & $14.751^{\pm 0.208}$ & $1.898^{\pm 0.080}$ & $2.768^{\pm 0.108}$ \\
 & w/o One-way attention & $\underline{1.354}^{\pm 0.026}$ & $\mathbf{1.361}^{\pm 0.011}$ & $\mathbf{1.329}^{\pm 0.022}$ & $14.863^{\pm 0.244}$ & $1.392^{\pm 0.089}$ & $5.066^{\pm 0.159}$ \\
\midrule
\multirow{4}{*}{Audio}
 & MixReact (Ours)  & $\mathbf{1.326}^{\pm 0.006}$ & $\mathbf{1.361}^{\pm 0.003}$ & $\underline{1.324}^{\pm 0.003}$ & $\underline{13.920}^{\pm 0.061}$ & $\underline{0.987}^{\pm 0.028}$ & $\underline{2.645}^{\pm 0.034}$    \\
 & w/o Discrete branch   & $1.616^{\pm 0.014}$ & $1.011^{\pm 0.038}$ & $0.816^{\pm 0.089}$ & $\mathbf{11.258}^{\pm 0.007}$ & $\mathbf{0.131}^{\pm 0.001}$ & $\mathbf{0.905}^{\pm 0.031}$ \\
 & w/o Continuous branch  & $1.417^{\pm 0.018}$ & $1.311^{\pm 0.027}$ & $1.285^{\pm 0.012}$ & $14.813^{\pm 0.0642}$ & $1.880^{\pm 0.096}$ & $2.796^{\pm 0.246}$ \\
 & w/o One-way attention & $\underline{1.362}^{\pm 0.009}$ & $\underline{1.363}^{\pm 0.012}$ & $\mathbf{1.327}^{\pm 0.014}$ & $15.160^{\pm 0.270}$ & $1.543^{\pm 0.150}$ & $5.330^{\pm 0.297}$ \\
\midrule
\multirow{4}{*}{Video}
 & MixReact (Ours)  &$\mathbf{0.706}^{\pm 0.003}$ & $\mathbf{1.382}^{\pm 0.003}$ & $\underline{1.244}^{\pm 0.003}$ & $20.908^{\pm 0.185}$ & $\underline{3.081}^{\pm 0.062}$ & $\underline{7.505}^{\pm 0.075}$     \\
 & w/o Discrete branch   & $0.831^{\pm 0.019}$ & $1.347^{\pm 0.015}$ & $1.101^{\pm 0.025}$ & $26.158^{\pm 0.144}$ & $5.502^{\pm 0.125}$ & $16.886^{\pm 0.201}$ \\
 & w/o Continuous branch  & $0.848^{\pm 0.012}$ & $1.353^{\pm 0.006}$ & $\mathbf{1.269}^{\pm 0.024}$ & $\mathbf{18.521}^{\pm 0.671}$ & $3.305^{\pm 0.084}$ & $\mathbf{4.659}^{\pm 0.218}$ \\
 & w/o One-way attention & $\underline{0.722}^{\pm 0.016}$ & $\underline{1.375}^{\pm 0.012}$ & $1.229^{\pm 0.021}$ & $\underline{20.249}^{\pm 1.009}$ & $\mathbf{3.040}^{\pm 0.121}$ & $9.601^{\pm 0.411}$ \\
\bottomrule
\end{tabular}%
}
\label{table:ablation}
\end{table*}
\begin{figure*}
    \centering
    \includegraphics[width=1.0\textwidth]{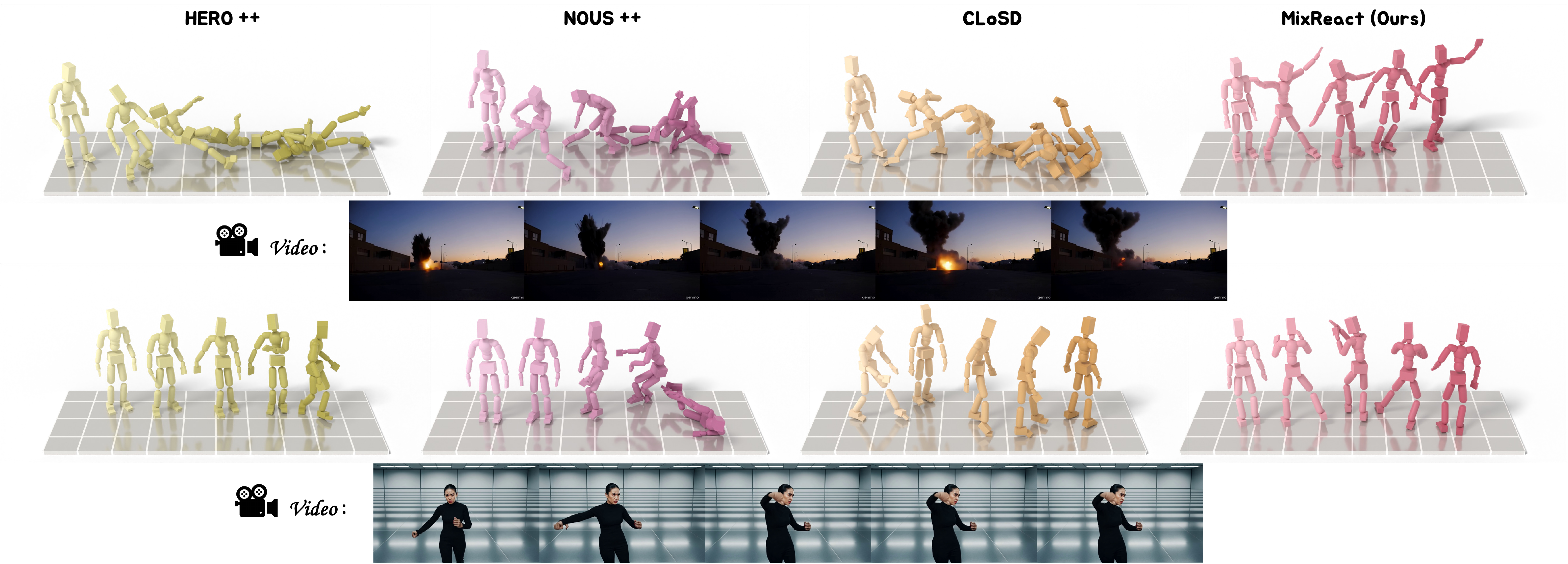}
    \caption{\textbf{Qualitative comparison of video-driven humanoid reaction control.} MixReact responds to the explosion with a startled retreat and to the striking action with a defensive posture. The baselines lose balance in the explosion example, while HERO++~\cite{yu2025hero,luo2023perpetual} and CLoSD~\cite{tevet2025closd} exhibit less pronounced defensive responses to the striking action.}
    \label{fig:comp_video}
\end{figure*}
\paragraph{Inference.}
As shown in Fig.~\ref{fig:overview}, for a rollout of $T$ frames, the discrete branch samples a complete sequence of $N=\lceil T/\kappa\rceil$ state tokens once. Starting from a fully masked sequence, it performs $R$ sampling rounds, drawing token proposals from categorical distributions and committing them at randomly selected unresolved positions until the sequence is complete. Committed tokens remain fixed, and the completed sequence provides intermediate-layer key--value caches as persistent global context.
Then, the continuous branch performs 10 denoising steps conditioned on semantic memory $C$, the cached discrete context, and the latest execution history $H_t^{d}$. At control step $t$, HAT maps the predicted next state $\hat{s}_{t+1}$ and history $H_t^{a}$ to an executable action $a_t$. Both histories are updated exclusively with simulator-returned states, closing the loop between state prediction and physical execution. Throughout the physics-based rollout, persistent discrete guidance supports consistency of the selected reaction across successive windows, while direct prediction of future states in continuous space bypasses the bottleneck of discrete representations.
\section{Experiments}
\label{sec:experiments}
\subsection{Experimental Settings}
\begin{figure*}
    \centering
    \includegraphics[width=1.0\textwidth]{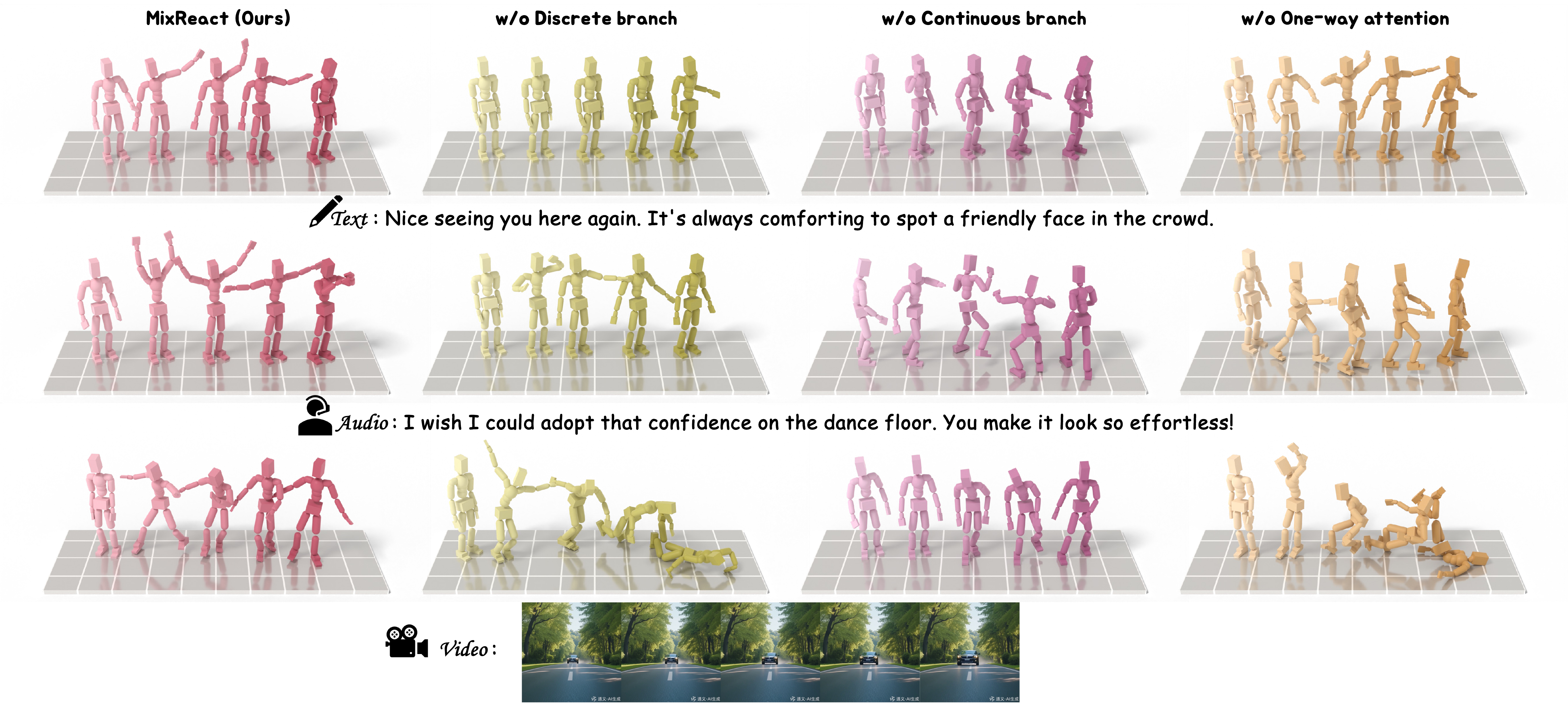}
    \caption{\textbf{Ablation study of MixReact.} Qualitative comparison of MixReact and its variants.}
    \label{fig:ablation_vis}
\end{figure*}
\paragraph{Datasets and evaluation metrics.}
We evaluate our model on ReactMotionNet~\cite{luo2026reactmotion} for text- and audio-driven reaction generation, and on ViMo~\cite{yu2025hero} for video-driven reaction generation. We follow the standard evaluation protocol for physics-based motion generation~\cite{yuan2023physdiff, tevet2025closd,wu2025uniphys}, with details provided in the supplementary material.

\subsection{Comparison}
\paragraph{Baseline methods.}
\text{ReactMotion}~\cite{luo2026reactmotion} generates text- and audio-driven listener reactions through preference-trained autoregressive modeling.
\text{HERO}~\cite{yu2025hero} generates video-driven reactions through intention-guided masked modeling, while NOUS~\cite{zhou2026nousvideodriven3dhuman} refines reactions with prototype-aligned visual features.
We construct ReactMotion++~\cite{luo2026reactmotion,luo2023perpetual}, HERO++~\cite{yu2025hero,luo2023perpetual}, and NOUS++~\cite{zhou2026nousvideodriven3dhuman,luo2023perpetual} by executing their trajectories with the PHC tracking policy~\cite{luo2023perpetual} in simulation.
We modify CLoSD~\cite{tevet2025closd}, which couples autoregressive diffusion planning with closed-loop tracking, to support text-, audio-, and video-driven reactions.

\paragraph{Quantitative results.}
Tabs.~\ref{table:compare_baseline_reactmotion} and~\ref{table:compare_baseline_vimo} show consistent improvements across all modalities. Among the evaluated methods, MixReact achieves the lowest FID, highest multimodality, diversity closest to Phys-GT, and lowest floating and skating errors, with competitive jerk. On ViMo~\cite{yu2025hero}, it reduces FID by 17.5\% and skating error by 34.2\% relative to NOUS++~\cite{zhou2026nousvideodriven3dhuman,luo2023perpetual}; for audio-driven reactions, it reduces floating error by 40.9\% over ReactMotion++~\cite{luo2026reactmotion,luo2023perpetual}. These gains reflect complementary discrete global guidance and continuous state prediction: persistent discrete context supports coherent, diverse reactions, while continuous prediction captures fine-grained state evolution beyond the decoding bottleneck of discrete representations. Unlike the ``++'' baselines, which track fixed kinematic references, MixReact conditions both future-state prediction and physics-based control on actual execution history. Compared with CLoSD~\cite{tevet2025closd}, which also incorporates simulation feedback, MixReact achieves greater conditional diversity and lower physical errors, further supporting the coupling of persistent discrete guidance with continuous adaptation.

\paragraph{Qualitative results.}
Figs.~\ref{fig:comp_text}, \ref{fig:comp_audio}, and~\ref{fig:comp_video} illustrate expressive reactions with coordination and dynamic balance: MixReact claps to a celebratory cue and dances to rhythmic audio. ReactMotion++~\cite{luo2026reactmotion,luo2023perpetual} loses balance in several examples, whereas CLoSD~\cite{tevet2025closd} produces comparatively restrained standing responses in the illustrated text/audio cases. Under the explosion cue, MixReact performs a startled retreat while maintaining balance, whereas HERO++~\cite{yu2025hero,luo2023perpetual}, NOUS++~\cite{zhou2026nousvideodriven3dhuman,luo2023perpetual}, and CLoSD~\cite{tevet2025closd} fall. These comparisons highlight the benefit of MixReact's complementary design: persistent discrete context preserves the reaction structure, while continuous denoising and history-aware closed-loop control adapt physics-based execution details to the humanoid's evolving state.

\subsection{Ablation Study}
Tab.~\ref{table:ablation} and Fig.~\ref{fig:ablation_vis} evaluate the complementary roles of the discrete branch, continuous diffusion branch, and one-way mixture attention. Implementation details of the ablation variants are provided in the supplementary material.

\paragraph{Discrete branch.}
Removing the discrete branch (Fig.~\ref{fig:ablation_vis}, second column) produces a restrained response and loss of balance in the video example. Across modalities, FID increases, while diversity and multimodality decrease, including a 38.4\% drop in audio multimodality. These results support discrete planning for selecting distinct reactions to ambiguous cues and maintaining global context across prediction windows. Although text/audio floating and skating errors decrease, the accompanying reduction in behavioral variety underscores the limitations of physical metrics alone.

\paragraph{Continuous branch.}
Removing continuous denoising (Fig.~\ref{fig:ablation_vis}, third column) reduces arm-gesture amplitude in the text and video examples, worsens FID across modalities—including a 20.1\% increase for video—and increases text/audio floating and skating errors. These findings support complementing discrete behavior selection with continuous state prediction, which captures fine-grained details of the predicted future humanoid states beyond the decoding bottleneck of discrete representations and adapts the selected behavior to actual execution through history conditioning across successive prediction windows.

\paragraph{One-way mixture attention.}
Replacing one-way mixture attention with bidirectional attention increases both FID and jerk across all modalities, indicating reduced fidelity and smoothness. The visual results in Fig.~\ref{fig:ablation_vis} (fourth column) further illustrate the suboptimal behavior of the bidirectional variant. These support our one-way design, which protects discrete modeling from interference by continuous denoising, preserving global reaction structure to provide consistent guidance for continuous state prediction and physically plausible execution across prediction windows.

\section{Conclusion}
\label{sec:conclusion}
In this paper, we introduce MixReact for physics-based reactive humanoid control from text, audio, and video cues. DisCo-MoT couples discrete guidance with continuous prediction through one-way mixture attention; HAT enables robust closed-loop control. Extensive experiments demonstrate improved fidelity, diversity, and physical consistency. We hope this work inspires further research in reaction generation and advances the development of humanoid agents that interact naturally with the world.
\section{LLM Usage Statement}

We used Claude and ChatGPT to assist with manuscript translation, drafting and organization, language and grammar editing, and LaTeX formatting, including related-work descriptions and the presentation of experimental findings. These tools also provided advisory feedback on research methodology, experimental design, and the interpretation of results. They were not used to generate synthetic datasets, develop theoretical models or conceptual frameworks, or formulate mathematical claims. All AI-assisted content was reviewed and revised by the authors. References and related-work descriptions were checked against the original publications, and technical descriptions and result interpretations were verified against the implementation and experimental records. We take full responsibility for the final manuscript and all associated claims and artifacts, including content produced with generative AI assistance.

\section{Acknowledgement}
This project was supported by Shanghai Local College Capacity Building Program (23010503100), NSFC (No.62406195, W2431046, W2431046), Shanghai Frontiers Science Center of Human-centered Artificial Intelligence (ShangHAI), MoE Key Laboratory of Intelligent Perception and Human-Machine Collaboration (ShanghaiTech University), the Shanghai Frontiers Science Center of Human-centered Artificial Intelligence, HPC Platform and Core Facility Platform of Computer Science and Communication of ShanghaiTech University and Shanghai Engineering Research Center of Intelligent Vision and Imaging.

{
    \small
    \bibliographystyle{ieeenat_fullname}
    \bibliography{main}
}

\clearpage
\setcounter{page}{1}
\maketitlesupplementary

This supplementary material provides further details and analyses of MixReact. We introduce the accompanying demo video (Sec.~\ref{demo}) and describe HAT's online distillation and comparison with its teacher (Sec.~\ref{HAT training}). Additional experiments cover fall-rate analysis, the user study protocol and results, an ablation comparing discrete and continuous global modeling, runtime and real-time performance, and qualitative examples of one-to-many reaction generation (Sec.~\ref{add experiment}). We detail state-based VQ-VAE~\cite{van2017neural} pretraining, modality encoders and conditioning, and the training, inference, and implementation settings of MixReact (Sec.~\ref{add training and infer}). We further describe the physics simulation, humanoid state representation, data collection, evaluation metrics and semantic assessment, and multimodal adaptation of CLoSD~\cite{tevet2025closd} (Sec.~\ref{add setup}). The ablation section specifies variant implementations and examines one-way mixture attention and the influence of dataset characteristics (Sec.~\ref{detail ablation}). We further analyze how persistent discrete guidance complements continuous diffusion for one-to-many generation (Sec.~\ref{analysis one2many}). Finally, we discuss current limitations and future directions for scene-aware interaction, real-world deployment, and joint multimodal conditioning (Sec.~\ref{limitations}).

\section{Demo video}\label{demo}
The accompanying demo video complements the static figures with detailed visualizations of the generated reactions, providing a more comprehensive view of our method’s performance during physical execution.

\section{Online distillation of HAT.}\label{HAT training}
To account for deviations arising from its own execution, we train HAT through online distillation based on DAgger~\cite{ross2011reduction} from a frozen PHC teacher~\cite{luo2023perpetual}. Given a reference motion, HAT predicts and executes $a_t=\pi(H_t^a,s_{t+1}^{\mathrm{ref}})$, conditioned on its observed execution history $H_t^a$ and the next reference state $s_{t+1}^{\mathrm{ref}}$. The reference state uses the same representation as DisCo-MoT predictions. At each student-visited state, the teacher receives a tracking observation $o_t^{\mathrm{T}}$ comprising the current simulated state and reference tracking information, and provides the target action $a_t^{\mathrm{T}}=\pi_{\mathrm{T}}(o_t^{\mathrm{T}})$. We minimize
\begin{equation}
\mathcal{L}_{\mathrm{distill}}
=
\mathbb{E}_{\mathcal{D}_{\pi}}
\left[
\frac{1}{d_a}
\left\|a_t-\operatorname{sg}(a_t^{\mathrm{T}})\right\|_2^2
\right],
\label{eq:controller_distillation}
\end{equation}
where $\mathcal{D}_{\pi}$ denotes samples collected through HAT rollouts, $d_a$ is the action dimension, and $\operatorname{sg}$ denotes stop-gradient. After distillation, HAT remains frozen during DisCo-MoT training and inference. At inference, the denormalized prediction $\hat{s}_{t+1}$ replaces $s_{t+1}^{\mathrm{ref}}$, while simulator feedback updates the execution histories for subsequent control and periodic state prediction.

\paragraph{Effectiveness of HAT.}
We compare HAT with its PHC teacher~\cite{luo2023perpetual} for executing reactions generated by DisCo-MoT (Table~\ref{table:hat}). To use DisCo-MoT's outputs directly, HAT takes the predicted 358-dimensional states as input and incorporates observed execution history to improve robustness during physical execution. To assess the effect of controller choice in the complete pipeline, we compare HAT and PHC using the same DisCo-MoT model and inference settings. Fall rate is defined as the percentage of executed trajectories in which the actual pelvis height falls below $0.5\,\mathrm{m}$ at any frame. HAT consistently reduces skating error and jerk across all three modalities, indicating improved contact consistency and motion smoothness. It also reduces fall rates from $0.819\%$ to $0.302\%$ for text, from $0.632\%$ to $0.374\%$ for audio, and from $18.286\%$ to $15.886\%$ for video. The substantially higher fall rates on ViMo~\cite{yu2025hero} are consistent with differences in motion distribution: its larger-amplitude reactions place greater demands on balance control than the conversational reactions in ReactMotionNet~\cite{luo2026reactmotion}. These results support HAT as a history-aware execution interface that directly accommodates DisCo-MoT predictions and enables smoother, more robust physical execution than its teacher.

\begin{table*}
\caption{\textbf{Comparison of HAT with its teacher.} HAT achieves lower skating errors, jerk, and fall rates across all three cue modalities, demonstrating smoother and more robust physical execution.}
\centering
\resizebox{\textwidth}{!}{%
\begin{tabular}{clccccccc}
\toprule
 &  & FID $\downarrow$ & Diversity $\rightarrow$ & MModality $\uparrow$ & Floating $[mm]$ $\downarrow$ & Skating $[mm]$ $\downarrow$ & Jerk $[mm/frame^3]$ $\downarrow$ & Fall rate$[\%]$ $\downarrow$ \\
\midrule
\multirow{2}{*}{Text}
 & HAT (Student)  & ${1.328}^{\pm 0.005}$ & ${1.356}^{\pm 0.003}$ & ${1.320}^{\pm 0.003}$ & ${13.827}^{\pm 0.086}$ & ${0.917}^{\pm 0.028}$ & ${2.626}^{\pm 0.036}$ & ${0.302}$  \\
 & PHC~\cite{luo2023perpetual} (Teacher) & ${1.342}^{\pm 0.006}$ & ${1.348}^{\pm 0.021}$ & ${1.302}^{\pm 0.013}$ & ${13.746}^{\pm 0.110}$ & $1.291^{\pm 0.067}$ & $3.837^{\pm 0.171}$ & ${0.819}$ \\
\midrule
\multirow{2}{*}{Audio}
 & HAT (Student)   & ${1.326}^{\pm 0.006}$ & ${1.361}^{\pm 0.003}$ & ${1.324}^{\pm 0.003}$ & ${13.920}^{\pm 0.061}$ & ${0.987}^{\pm 0.028}$ & ${2.645}^{\pm 0.034}$ & ${0.374}$  \\
 & PHC~\cite{luo2023perpetual} (Teacher) & ${1.348}^{\pm 0.008}$ & ${1.350}^{\pm 0.011}$ & ${1.306}^{\pm 0.017}$ & ${13.802}^{\pm 0.141}$ & $1.335^{\pm 0.109}$ & $3.803^{\pm 0.071}$ & $0.632$\\
\midrule
\multirow{2}{*}{Video}
 & HAT (Student)  &${0.706}^{\pm 0.003}$ & ${1.382}^{\pm 0.003}$ & ${1.244}^{\pm 0.003}$ & $20.908^{\pm 0.185}$ & ${3.081}^{\pm 0.062}$ & ${7.505}^{\pm 0.075}$  & ${15.886}$  \\
 & PHC~\cite{luo2023perpetual} (Teacher) & ${0.703}^{\pm 0.007}$ & ${1.378}^{\pm 0.006}$ & ${1.249}^{\pm 0.021}$ & $20.742^{\pm 0.086}$ & $3.780^{\pm 0.261}$ & $11.919^{\pm 0.292}$ & $18.286$\\
\bottomrule
\end{tabular}%
}
\label{table:hat}
\end{table*}

\section{Additional Experimental Results}\label{add experiment}
\subsection{Fall rates.}
\begin{figure}[t]
    \centering
    \includegraphics[width=1.0\columnwidth]{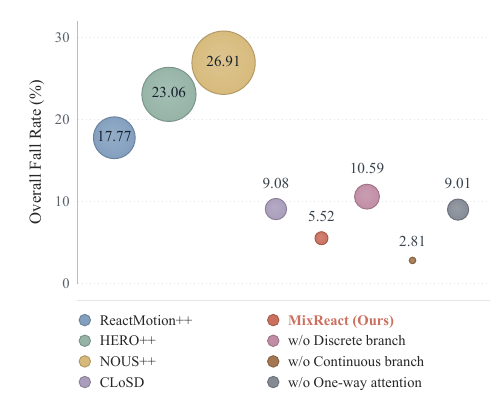}
    \caption{\textbf{Overall fall rates of MixReact, baselines, and ablated variants.} A trajectory is counted as a fall if its simulated pelvis height
  drops below 0.5 m at any frame. We compute the overall rate by averaging equally over each method’s supported modalities and
  then across random seeds. Colors identify methods; vertical position and bubble radius encode the mean fall rate (\%). Lower values
  indicate fewer falls.}
    \label{fig:fall_rate}
\end{figure}

Following the protocol in Sec.~\ref{add setup}, we compute fall rates using all simulated trajectories, averaging equally across each method's supported cue modalities and then across 10 random seeds.
As shown in Fig.~\ref{fig:fall_rate}, MixReact achieves an overall fall rate of 5.52\%, outperforming ReactMotion++ (17.77\%), HERO++ (23.06\%), NOUS++ (26.91\%), and CLoSD (9.08\%).
Removing the continuous branch further reduces the fall rate to 2.81\%.
This reduction is consistent with attenuated motion detail and amplitude under codebook and reconstruction constraints.
In Fig.~6 of the main paper, this variant exhibits more restrained arm gestures and whole-body movements in the text and video examples, respectively.
Tab.~3 provides complementary evidence: the variant has lower video jerk (4.659 versus 7.505), but worse FID across all three modalities, including 0.848 versus 0.706 for video.
These observations suggest that its lower fall rate partly reflects reduced execution demands at the expense of reaction fidelity and expressiveness.
Together with the results in Tab.~1 and~2 of the main paper, these findings demonstrate MixReact's advantages in balancing diverse, high-fidelity reaction generation with robust physical execution.

\subsection{User Study}
\begin{figure*}
    \centering
    \includegraphics[width=1.0\textwidth]{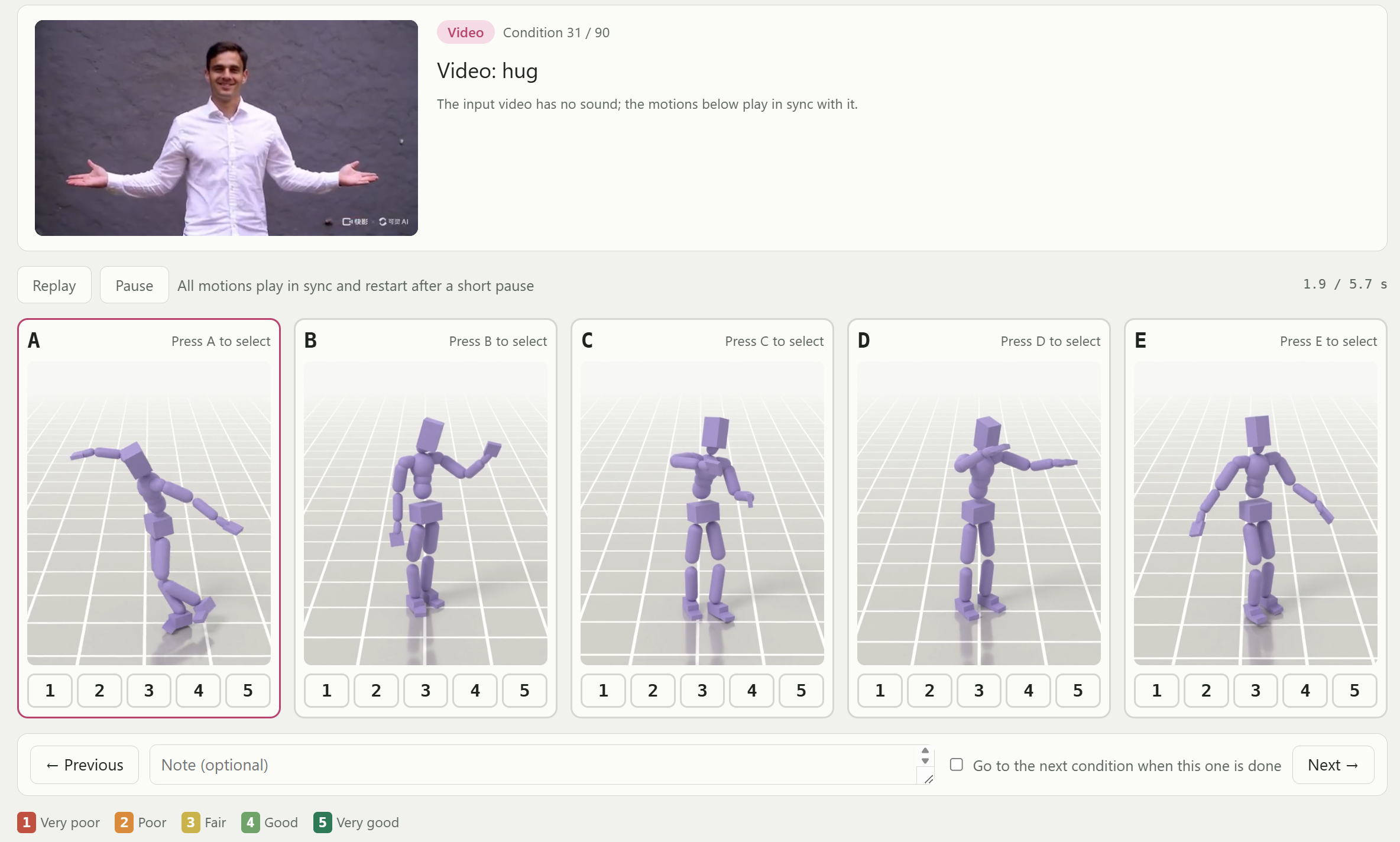}
    \caption{\textbf{User study interface.} Participants view the input cue alongside generated reactions from different methods and rate each reaction on a five-point scale, ranging from 1 (very poor) to 5 (very good).}
    \label{fig:user study page}
\end{figure*}

\paragraph{User study setup.}
We organize two separate rating rounds for baseline comparisons and ablation studies, both evaluated by the same 20 participants. In each round, we randomly select 30 cues per modality from the test set, yielding 90 cues spanning text, audio, and video. For each cue, reactions are presented together in randomized order under anonymous method labels, with synchronized playback to facilitate comparison (Fig.~\ref{fig:user study page}). Participants assign each reaction a single score from 1 (very poor) to 5 (very good), jointly considering reaction quality and semantic correspondence to the input cue. Each baseline is evaluated only on its supported modalities. We report average participant ratings separately for each round; MixReact's scores differ between the two rounds because the user studies are conducted separately, with cues randomly sampled from the test set for each round.

\begin{figure*}
    \centering
    \includegraphics[width=1.0\textwidth]{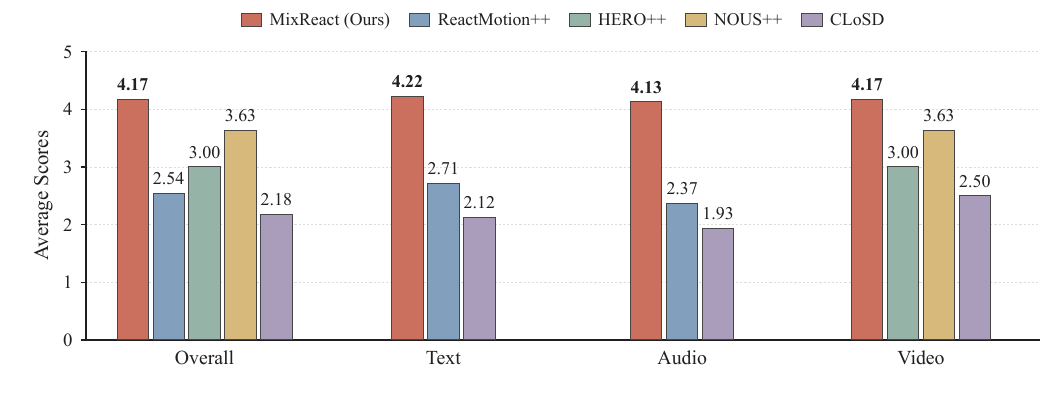}
    \caption{\textbf{User study comparison with baselines.} MixReact receives the highest average ratings across all three cue modalities.}
    \label{fig:user study comp}
\end{figure*}

\paragraph{User study results.}
In the baseline comparison (Fig.~\ref{fig:user study comp}), MixReact receives the highest ratings across all three modalities: 4.22 for text, 4.13 for audio, and 4.17 for video. Its advantages are particularly pronounced for text and audio, where the strongest competing baseline, ReactMotion++~\cite{luo2026reactmotion}, scores 2.71 and 2.37, respectively. For video, MixReact also outperforms NOUS++~\cite{zhou2026nousvideodriven3dhuman}, the highest-rated baseline at 3.63. The ablation (Fig.~\ref{fig:user study ablation}) further supports the contributions of our design components. Removing the discrete branch produces the largest decline across all modalities, reducing the overall rating from 3.60 to 2.07 and highlighting the importance of discrete behavioral guidance. Replacing one-way mixture attention with bidirectional attention lowers the overall rating to 2.87, supporting the benefit of protecting discrete context from continuous denoising. Removing the continuous branch reduces ratings for audio and video, lowering the overall score to 3.37 and supporting its contribution to detailed state prediction. These results demonstrate the perceptual advantages of MixReact and support the complementary roles of its components in generating high-quality, cue-appropriate reactions.

\subsection{Discrete versus Continuous Global Modeling}

\begin{table}[t]
\centering
\caption{\textbf{Discrete versus continuous global modeling.}
We replace the discrete global branch with Continuous Global Diffusion (CGD), which directly generates complete reaction sequences of 358-dimensional raw states, while keeping the local autoregressive diffusion branch unchanged.}
\label{table:discrete_global}
\small
\setlength{\tabcolsep}{3pt}
\resizebox{\columnwidth}{!}{%
\begin{tabular}{@{}clccc@{}}
\toprule
 & Method & FID $\downarrow$ & Diversity $\rightarrow$ & MModality $\uparrow$ \\
\midrule
\multirow{2}{*}{Text}
 & MixReact (Ours)
 & $1.328^{\pm 0.005}$
 & $1.356^{\pm 0.003}$
 & $1.320^{\pm 0.003}$ \\
 & Ours w/ CGD
 & $1.301^{\pm 0.002}$
 & $1.316^{\pm 0.004}$
 & $1.040^{\pm 0.011}$ \\
\midrule
\multirow{2}{*}{Audio}
 & MixReact (Ours)
 & $1.326^{\pm 0.006}$
 & $1.361^{\pm 0.003}$
 & $1.324^{\pm 0.003}$ \\
 & Ours w/ CGD
 & $1.314^{\pm 0.002}$
 & $1.295^{\pm 0.004}$
 & $1.086^{\pm 0.012}$ \\
\midrule
\multirow{2}{*}{Video}
 & MixReact (Ours)
 & $0.706^{\pm 0.003}$
 & $1.382^{\pm 0.003}$
 & $1.244^{\pm 0.003}$ \\
 & Ours w/ CGD
 & $0.768^{\pm 0.002}$
 & $1.362^{\pm 0.003}$
 & $1.103^{\pm 0.013}$ \\
\bottomrule
\end{tabular}%
}
\end{table}
To assess the contribution of discrete global modeling, we replace the discrete global branch of DisCo-MoT with Continuous Global Diffusion (CGD), while keeping the local autoregressive diffusion branch unchanged.
Both variants generate complete global plans, have the same parameter count, and use the same HAT controller, allowing us to evaluate the benefit of discrete modeling beyond the availability of global context.
As shown in Table~\ref{table:discrete_global}, MixReact consistently achieves higher Diversity and MModality across all three modalities.
In particular, MModality increases from 1.040 to 1.320 for text, from 1.086 to 1.324 for audio, and from 1.103 to 1.244 for video.
These gains accompany comparable text/audio FID and improved video FID, from 0.768 to 0.706.
These results support discrete global modeling as more effective at capturing one-to-many reactions while maintaining competitive distributional fidelity.
\subsection{Runtime and Real-time Performance.}
We evaluate MixReact on a single NVIDIA RTX 3090 GPU with FP32 precision.
The discrete branch samples a complete reaction plan and constructs its intermediate-layer key--value caches in 52.47\,ms on average. This computation is performed once per reaction, and the resulting caches are reused throughout execution. Conditioned on the cached context, the continuous branch predicts a window of 60 future states in 79.49\,ms, corresponding to 12.58 prediction windows per second, or an equivalent generation throughput of 754.8 states per second. At the nominal control frequency of 30\,Hz, each window covers 2\,s of simulated motion. A new window is generated every 60 control steps, yielding a nominal replanning frequency of 0.5\,Hz in simulation time. The initial DisCo-MoT prediction, including discrete-plan preparation and generation of the first future-state window, takes 131.90\,ms on average.

At each control step, HAT conditions on the latest 16 observed states and the predicted next state to produce an executable action. Its standalone inference latency is 3.39\,ms on average, corresponding to a throughput of 294.8 actions per second, well above the nominal 30\,Hz control rate.

To assess closed-loop execution speed, we additionally benchmark the complete closed loop, including DisCo-MoT state prediction, HAT inference, physics simulation, and simulator state feedback. For each modality, we conduct five runs of 100 control steps each. The measured wall-clock throughputs are 34.56, 34.63, and 34.35 control steps per second for text, audio, and video, respectively, with an aggregate throughput of 34.51 steps per second. Relative to the nominal 30\,Hz control rate, this corresponds to approximately $1.15\times$ real-time execution on average.

\subsection{One-to-many Reaction Generation.}
To further illustrate MixReact's capacity for one-to-many generation, Fig.~\ref{fig:one2many} presents three reactions to each fixed text, audio, or video cue. The generated reactions exhibit distinct combinations of gestures and postural changes under the same conditioning input. For example, upon receiving the remorseful audio cue, the humanoid may bring its hands toward its face to convey empathetic concern, raise a hand to signal reassurance, or lean forward with outward hand gestures to invite the speaker to elaborate. These alternatives differ in their movement patterns while remaining plausible responses to the shared cue. Across all three modalities, the examples illustrate MixReact's ability to generate diverse behavioral realizations of a single cue and execute them in physics-based simulation.

\section{Additional Training and Inference Details}\label{add training and infer}

\subsection{Training Details of VQ-VAE}
\label{subsec:vqvae_training}

We pretrain a VQ-VAE~\cite{van2017neural} on humanoid state trajectories to learn the discrete vocabulary used by DisCo-MoT. After excluding sequences shorter than 32 frames, the pretraining corpus comprises 20,188 trajectories from the AMASS~\cite{mahmood2019amass} training set, with no trajectory overlap with the test sets used to evaluate MixReact. Each state channel is standardized using statistics computed over the retained frames. We retain the original trajectory lengths and extract 32-frame windows with a stride of one, without temporal resampling.

Following T2M-GPT~\cite{zhang2023generating}, the temporal convolutional encoder has a hidden width of 512 and comprises two downsampling stages. Each stage consists of a stride-2 convolution followed by three residual blocks with dilation rates of 1, 3, and 9. The resulting temporal downsampling factor is $\kappa=4$, yielding eight tokens per 32-frame window. Encoder features are quantized through nearest-neighbor assignment to a single codebook containing $K=512$ entries, each of dimension 512. The decoder reconstructs normalized states through dilated residual blocks, nearest-neighbor upsampling, and temporal convolutions.

Let $s$ and $\widehat{s}$ denote the normalized input and reconstructed state windows, respectively, and let $f$ and $f_q$ denote the encoder features and their quantized counterparts. The training objective is
\begin{equation}
\begin{aligned}
\mathcal{L}_{\mathrm{VQ}}
=\operatorname{SL}_1(\widehat{s},s) 
+ \operatorname{MSE}\!\left(f,\operatorname{sg}(f_q)\right)
+ \lambda\,\operatorname{SL}_1\!\left(
\Delta\widehat{s},\Delta s
\right),
\end{aligned}
\end{equation}
where $\operatorname{SL}_1$ denotes the Smooth L1 loss, $\operatorname{sg}$ denotes stop-gradient, $\Delta s_t=s_{t+1}-s_t$, and $\lambda$ is a hyperparameter controlling the weight of the temporal-difference loss. We set $\lambda=0.5$. The codebook is updated using exponential moving averages with a decay of 0.99. Entries with an EMA assignment count below one are reinitialized from encoder features in the current batch. The reconstruction term encourages accurate state recovery, the commitment term constrains encoder features to remain close to their assigned code vectors, and the temporal-difference term encourages preservation of frame-to-frame state changes. A straight-through estimator propagates reconstruction gradients through the quantization operation.

We optimize the encoder and decoder for 100 epochs, corresponding to 723,800 optimization steps, with a batch size of 512. We use AdamW~\cite{loshchilov2017decoupled} with a constant learning rate of $2\times10^{-4}$, $(\beta_1,\beta_2)=(0.9,0.99)$, and zero weight decay. The global gradient norm is clipped to 1.0.
After pretraining, the tokenizer is frozen and used to encode complete trajectories into discrete state tokens that supervise the discrete branch. During MixReact inference, the discrete Transformer samples a complete reaction plan and caches keys and values from its intermediate layers to guide continuous state prediction. The continuous branch predicts future states directly; neither the VQ-VAE encoder nor its decoder is invoked during standard online generation.

\subsection{Details of MixReact}

\paragraph{Modality encoders and conditioning.}
We extract cue features using frozen pretrained encoders. For text, we retain 768-dimensional token features from the text Transformer of CLIP ViT-L/14@336px~\cite{radford2021learningtransferablevisualmodels}. The audio head of AudioCLIP~\cite{guzhov2022audioclip} extracts a descriptor from mono audio resampled to 44.1\,kHz and center-cropped or zero-padded to 5\,s. For video, TC-CLIP~\cite{kim2024leveraging}, built on ViT-B/16~\cite{dosovitskiy2021imageworth16x16words}, jointly encodes 16 uniformly sampled frames resized to $224\times224$, yielding 512-dimensional frame-level features. MCA maps these features through modality-specific adapters of width 512 and a shared projection to construct the 384-dimensional semantic memory $C$. Each sample uses one cue modality, with padding and inactive modality slots masked out. MCA is optimized exclusively by the discrete masked-modeling objective, while the continuous branch accesses $C$ with stopped gradients.

\paragraph{Training details of DisCo-MoT.}
For each cue--reaction pair, a frozen modality-specific encoder extracts cue features, which MCA maps to semantic memory $C$. Both branches access $C$ through condition cross-attention. Complete state trajectories are tokenized offline by the frozen VQ-VAE~\cite{van2017neural}, using the normalization established during its pretraining.

Each epoch consists of two sequential optimization passes. In the discrete pass, tokens are independently masked with probability 0.5, and the discrete branch predicts their original identities conditioned on the visible tokens and $C$. Cross-entropy loss over masked positions updates the discrete branch and MCA.

In the continuous pass, we extract 30-frame history windows and up to 60 future frames from the original trajectories. Each window retains its source trajectory's complete ground-truth token sequence $Z$. A separate forward pass through the discrete branch processes the unmasked $Z$ and the same semantic memory $C$, without gradient computation, to construct intermediate-layer key--value caches. The continuous branch receives Gaussian-corrupted future states and accesses these caches through one-way mixture attention. Within this attention mechanism, continuous queries attend jointly to discrete and continuous keys and values, while discrete queries are restricted to discrete tokens. The continuous branch also attends separately to the clean history and $C$, and predicts normalized clean future states using a mean squared error loss over valid frames. Both the discrete caches and $C$ are detached, so this pass updates only the continuous branch.

\paragraph{Inference details.}
Given a cue, MCA constructs $C$, and the discrete branch starts from a fully masked token sequence. Over eight sampling rounds, it commits sampled tokens at randomly selected unresolved positions while retaining previously committed tokens. The completed sequence $\widehat{Z}$ undergoes an additional forward pass to construct key--value caches, which remain fixed throughout execution. We initialize the humanoid in a stationary standing pose and repeat its observed initial state to populate the history buffers. Conditioned on $C$, the cached discrete context, and the normalized 30-frame execution history, the continuous branch generates a window of up to 60 future states from Gaussian noise through 10 denoising steps. These predictions are denormalized and supplied sequentially to HAT. At each control step, HAT combines the corresponding predicted next state with the latest 16 observed states to generate an executable action. Simulator feedback updates both history buffers, and a new future window is generated every 60 control steps while reusing the same discrete context. Thus, subsequent predictions and actions depend on the states actually reached during execution.

\paragraph{DisCo-MoT Implementation details.}
We train DisCo-MoT on three NVIDIA RTX 3090 GPUs with 24\,GB memory each. Both branches use a global batch size of 3,072 (1,024 per GPU), comprising cue--motion records for discrete training and state windows for continuous training. Separate AdamW optimizers~\cite{loshchilov2017decoupled} use constant learning rates of $2\times10^{-4}$ and $10^{-4}$ and weight decay of 0.01 and 0 for the discrete and continuous branches, respectively, with $(\beta_1,\beta_2)=(0.9,0.999)$. We train DisCo-MoT for a total of 150 epochs, corresponding to 1,650 and 11,930 optimizer updates for the discrete and continuous branches, respectively. The continuous branch performs 10 DDPM denoising steps with a cosine noise schedule~\cite{nichol2021improved}. We further apply classifier-free guidance~\cite{ho2022classifier}, where cue conditioning is retained during discrete training and dropped with probability 0.1 during continuous training, and a guidance scale of 2.5 is used during diffusion sampling.

\paragraph{HAT implementation details.}
We train HAT on eight NVIDIA RTX 4090 GPUs with 24\,GB memory each. Each GPU runs 6,912 parallel simulation environments, yielding a global batch size of 55,296 online samples per update. HAT comprises four Transformer blocks with a hidden dimension of 512, eight attention heads, and dropout of 0.1. It takes a 16-frame history of 358-dimensional states and the next reference state as inputs and predicts a 69-dimensional action. Online distillation~\cite{ross2011reduction} uses a frozen PHC teacher~\cite{luo2023perpetual} and 20,188 reference motions from AMASS~\cite{mahmood2019amass} sampled at 30\,FPS, with no reference-motion overlap with our MixReact evaluation test sets. At each control step, the teacher provides action supervision at the states visited by HAT, while the simulator executes HAT's own predicted actions. We minimize the action mean squared error for 212,000 optimizer updates using AdamW~\cite{loshchilov2017decoupled} with a constant learning rate of $10^{-4}$, $(\beta_1,\beta_2)=(0.9,0.999)$, and weight decay of $2\times10^{-5}$. Bias, normalization, positional, and query parameters are exempt from weight decay, and the global gradient norm is clipped to 1.0. After distillation, HAT remains frozen during DisCo-MoT training and inference.
\begin{figure*}
    \centering
    \includegraphics[width=1.0\textwidth]{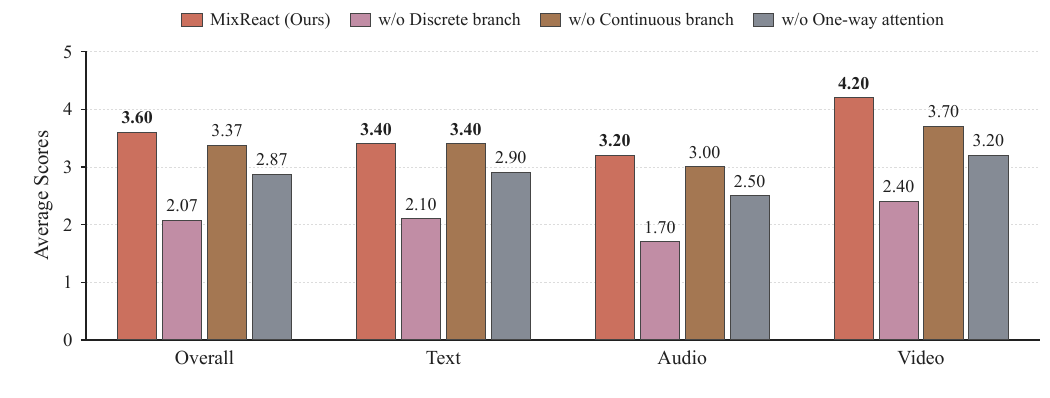}
    \caption{\textbf{User study of ablation variants.} MixReact achieves the highest overall user rating, with the largest decline observed when the discrete branch is removed across all three cue modalities, highlighting the importance of discrete guidance.}
    \label{fig:user study ablation}
\end{figure*}
\begin{figure*}
    \centering
    \includegraphics[width=1.0\textwidth]{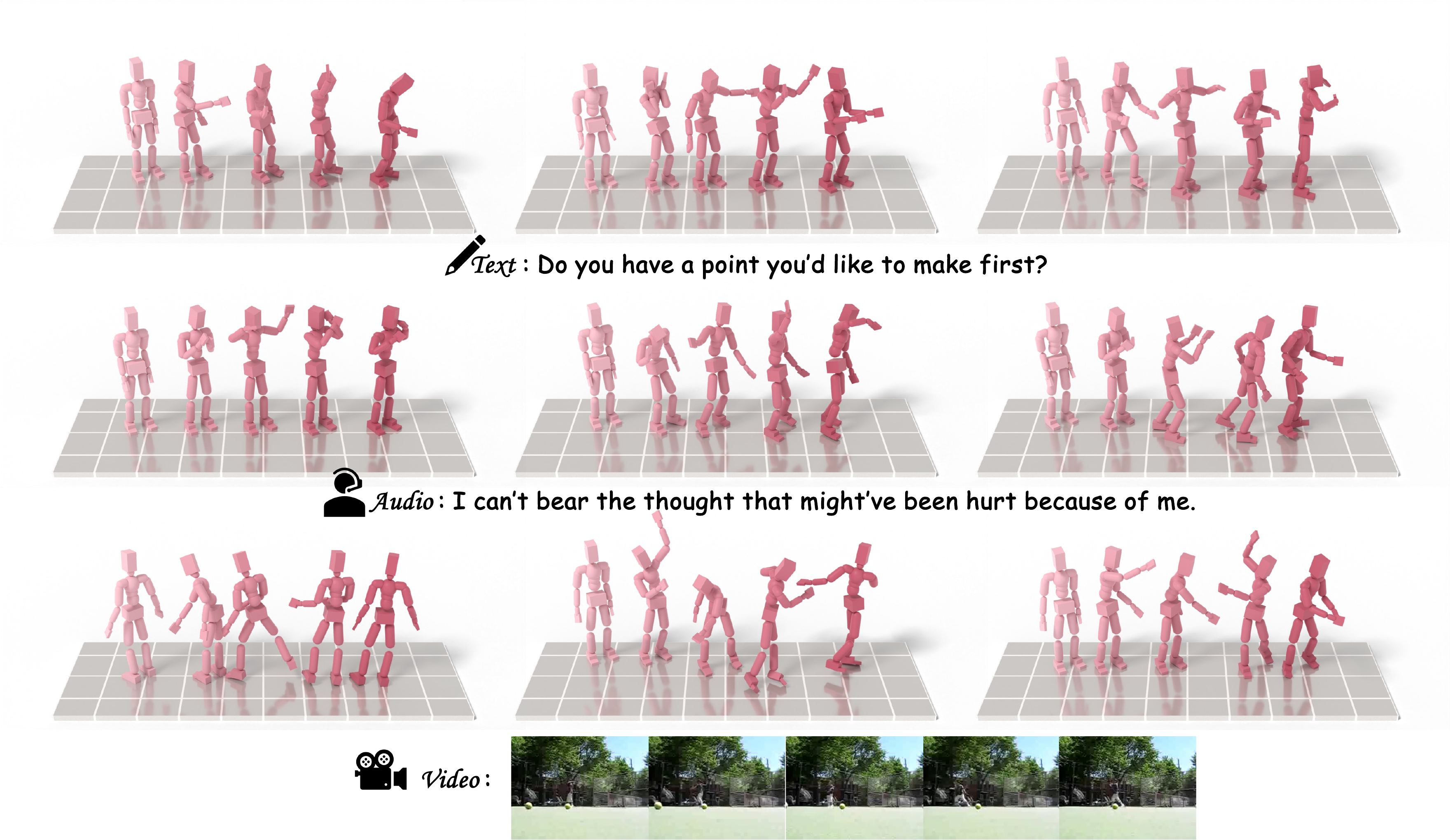}
    \caption{\textbf{One-to-many reaction generation.} Each row shows three distinct reactions generated by MixReact from the same text, audio, or video cue and executed in physics-based simulation. Colors progress from light to dark to indicate temporal order.}
    \label{fig:one2many}
\end{figure*}
\section{Additional Experimental Setup Details}\label{add setup}
\paragraph{Physics simulation setup.}
We use Isaac Gym~\cite{makoviychuk2021isaac} to simulate a humanoid with an SMPL-like~\cite{loper2015smpl} skeleton comprising 24 joints and 69 actuators. Each actuator is driven by a proportional-derivative (PD) controller. At time step $t$, the policy produces an action $\bm{a}_t\in\mathbb{R}^{69}$ specifying the target joint positions. The simulator evolves the current state $\bm{s}_t\in\mathbb{R}^{358}$ under this action according to the transition dynamics $\bm{s}_{t+1}\sim p(\bm{s}_{t+1}\mid\bm{s}_t,\bm{a}_t)$.

\paragraph{Humanoid state representation.}
Taking the pelvis as the root joint, we represent the humanoid state as $\bm{s}=(h,\bm{p},\bm{r},\bm{v},\bm{\omega})$. The scalar $h\in\mathbb{R}$ specifies the root height above the ground plane, while $\bm{p}\in\mathbb{R}^{23\times3}$ contains the positions of the non-root joints expressed in the local root-heading frame. The local joint orientations are encoded in $\bm{r}\in\mathbb{R}^{24\times6}$ using a continuous 6D rotation representation~\cite{zhou2019continuity}. The remaining components, $\bm{v}\in\mathbb{R}^{24\times3}$ and $\bm{\omega}\in\mathbb{R}^{24\times3}$, describe the local linear and angular velocities of all joints, respectively.
\paragraph{Data collection.}
We construct cue-associated trajectories from ReactMotionNet~\cite{luo2026reactmotion} for text and audio cues and ViMo~\cite{yu2025hero} for video cues. Reference motions are tracked in simulation using PHC~\cite{luo2023perpetual}, yielding humanoid state sequences and corresponding control actions. We retain only trajectories satisfying PHC's tracking success criteria to obtain high-quality, physically plausible expert demonstrations. Each retained state--action sequence is linked to its original cue annotations through motion identifiers, forming sequence-level state--action--condition triplets. We preserve the original dataset splits and retain training pairs with successfully tracked trajectories. The resulting training corpus comprises 29,615 ReactMotionNet records and 2,800 ViMo pairs, spanning 2,477 distinct trajectories and 421,957 state frames. Within this corpus, ReactMotionNet explicitly represents one-to-many cue--reaction associations by pairing speaker utterances with candidate listener motions annotated with graded appropriateness. We retain its Gold and Silver annotations, obtaining 29,515 distinct cue--motion pairs across 6,624 cue groups, with an average of 4.46 reference reactions per cue. Each record provides textual and auditory conditions associated with the same trajectory, and both modalities share these one-to-many associations. For video conditioning, we use ViMo, which provides one annotated reaction per video despite the underlying ambiguity of visual reaction generation: a friendly approach, for example, may elicit waving, handshaking, or hugging. Accordingly, ReactMotionNet explicitly supplies multiple reference reactions per cue, whereas ViMo provides a single reference for a task that admits diverse plausible reactions. Physical evaluation uses the same 696 test-set conditions and fixed Gold references from ReactMotionNet for both text- and audio-driven reactions, and 700 test-set conditions from ViMo for video-driven reactions.

\paragraph{Evaluation metrics.}
We evaluate reaction distribution quality, diversity, and physical consistency on simulated executions, retaining all valid frames, including those following a fall. To avoid leaking ground-truth sequence lengths and ensure fair comparisons, all methods generate fixed-length sequences of 180 frames (6 seconds at 30\,Hz), corresponding to the average trajectory length in the training set. FID, Diversity, and MModality are computed from features extracted by a frozen state encoder trained following the evaluator training protocol adopted in HERO~\cite{yu2025hero}. \emph{Fr\'echet Inception Distance (FID)} measures the distributional discrepancy between generated and Phys-GT reactions. \emph{Diversity} measures the average feature distance between randomly sampled motion pairs and is assessed relative to reference diversity, whereas \emph{MultiModality (MModality)} measures pairwise feature distances between reactions generated from the same cue. Physical metrics are computed from simulated rigid-body positions. \emph{Floating} measures the mean positive excess of the lowest rigid-body origin above a 5\,mm ground-clearance threshold. \emph{Skating} measures the average horizontal displacement of the toe rigid bodies over adjacent-frame intervals in which the same toe remains at or below 50\,mm in both frames. \emph{Jerk} measures the mean norm of third-order finite differences of rigid-body positions. \emph{Fall rate} is the percentage of trajectories whose actual pelvis height falls below $0.5\,\mathrm{m}$ at any frame, with each trajectory counted at most once under this height-based criterion.
We evaluate each method using $n=10$ different random seeds and report mean scores with 95\% normal-approximation confidence intervals:
\begin{equation}
\bar{m}\pm 1.96\frac{s}{\sqrt{n}},
\end{equation}
where $\bar{m}$ and $s$ denote the mean and sample standard deviation across the evaluation runs.

\paragraph{Evaluation of semantic appropriateness.}
We assess cue--reaction semantic correspondence through the qualitative results in Figs.~3--6 of the main paper and the accompanying demo video, which present generated reactions alongside their input cues. We further conduct a user study in which 20 participants view each cue and its generated reactions, assigning ratings from 1 to 5 based jointly on reaction quality and semantic appropriateness to the cue.

\paragraph{Adaptation of CLoSD.}
We adapt CLoSD~\cite{tevet2025closd} to support text-, audio-, and video-driven reactions within a shared model. We replace DiP's text-conditioning interface with trainable modality-specific adapters and a shared projection, using features extracted by the same frozen encoders as MixReact. The resulting conditioning tokens are supplied through cross-attention with padding masks, with one active modality per sample. We jointly train DiP and the adapters from scratch on ReactMotionNet~\cite{luo2026reactmotion} and ViMo~\cite{yu2025hero} cue--motion pairs, retaining the original 263-dimensional HumanML3D representation~\cite{guo2022generating} and diffusion objective. We further fine-tune CLoSD's PHC-based tracking controller~\cite{luo2023perpetual} using its native training pipeline. Generated motions are converted into reference trajectories through CLoSD's native representation pipeline and executed by the fine-tuned controller, while actual simulation history provides the prefix for subsequent predictions. Apart from the multimodal conditioning interface and reaction training data, we follow CLoSD's official training and inference implementation, retaining its original hyperparameters and model configurations.

\section{More Details of Ablations}\label{detail ablation}

\paragraph{Ablation implementation details.}
We retain the same data splits, modality features, and frozen HAT controller across ablation variants. \emph{w/o Discrete branch} removes the discrete Transformer, masked-token objective, and discrete key--value context. The continuous branch and MCA are trained jointly from scratch, with future states conditioned on the cue and execution history. \emph{w/o Continuous branch} reuses the trained discrete generator of our full model and decodes its sampled token sequence into state references through the frozen VQ-VAE, without additional training. These references remain fixed throughout execution, while HAT continues to use observed history for closed-loop control. \emph{w/o One-way mixture attention} retains both branches but allows their queries to attend jointly to discrete and continuous keys and values, removes cross-branch stop-gradients, and jointly optimizes both branches. Two aligned forward passes compute masked-token cross-entropy and clean-state reconstruction loss, respectively. The former uses masked ground-truth tokens, whereas the latter uses complete, unmasked tokens as discrete context. Both passes share the same noisy future window and history. Both optimizers are updated during the full-sequence and window-based training stages. At inference, sampled token identities remain fixed, but both branches recompute their features at every denoising step because the discrete features now depend on the evolving continuous states.

\paragraph{Further analysis of one-way mixture attention.}
The bidirectional variant allows both branches to attend to the combined discrete and continuous keys and values and jointly optimizes them without cross-branch stop-gradients. Although the sampled token identities remain fixed during inference, their intermediate representations depend on the evolving noisy state windows. During training, the diffusion objective also updates the discrete branch, exposing global behavioral representations to local reconstruction gradients. Our one-way design restricts discrete queries to discrete keys and values and blocks diffusion gradients through the discrete caches and semantic memory. This preserves the independence of full-sequence behavioral modeling while allowing discrete context to guide continuous state prediction.

Tab.~3 of the main paper supports this design: bidirectional coupling yields higher FID and jerk across all three modalities, with jerk approximately doubling for text and audio. It also increases floating and skating errors for text- and audio-driven reactions, indicating poorer contact consistency. The visual results provide complementary evidence: in the video example, the bidirectional variant loses balance and falls, whereas MixReact maintains balance throughout the reaction (Fig.~6 of the main paper, fourth and first columns, respectively). The user study further supports these findings (Fig.~\ref{fig:user study ablation}). Replacing one-way mixture attention with bidirectional coupling lowers the overall rating from 3.60 to 2.87, with decreases across all three modalities and the largest decline for video, from 4.20 to 3.20. These ratings reflect participants' joint assessment of reaction quality and cue correspondence, showing that the benefits extend to perceived reaction quality. Collectively, the quantitative, visual, and perceptual results support protecting global behavioral representations from interference by local continuous denoising, so that fine-grained state prediction remains guided by a consistent representation of the whole reaction.

\paragraph{Influence of dataset characteristics.}
The different ablation trends are consistent with differences in the cue--reaction associations represented by the two datasets. ReactMotionNet~\cite{luo2026reactmotion} pairs identical text or audio cues with multiple behaviorally distinct reactions, introducing heterogeneous alternatives under the same condition. As shown in Tab.~3 of the main paper, the continuous-only variant exhibits higher FID and lower multimodality than the discrete-only variant, consistent with a greater tendency toward behavioral collapse under explicit one-to-many supervision. Discrete modeling provides an explicit mechanism for selecting among these alternatives and preserving the selected reaction throughout generation. In contrast, ViMo~\cite{yu2025hero} provides a single annotated reaction per video, with related scenarios represented by similar but distinct visual cues. Consequently, its annotations do not explicitly associate each individual cue with multiple heterogeneous targets. On this dataset, the continuous-only variant achieves a slightly lower FID than the discrete-only variant, suggesting a benefit from predicting states directly rather than reconstructing them from discrete tokens. This observation is consistent with the continuous branch bypassing quantization-induced reconstruction errors while retaining history-conditioned adaptation during execution. These dataset-dependent trends motivate the complementary design of DisCo-MoT: discrete context supports distinct behavioral choices, while continuous prediction preserves fine-grained state information and adapts execution details. Their combination achieves the lowest FID among the evaluated variants across all three cue modalities while maintaining competitive physical consistency.

\section{Analysis of One-to-Many Generation}\label{analysis one2many}

Diffusion models can represent multimodal conditional distributions~\cite{ho2020denoising,tevet2022human}, and continuous representations do not inherently lead to averaged behaviors. Our setting nevertheless poses a distinct challenge: everyday cues leave the intended reaction unspecified and can elicit behaviorally distinct responses~\cite{yu2025hero,luo2026reactmotion}. For example, ``Congratulations!'' may prompt a bow, a handshake, or a celebratory jump. These alternatives differ in their overall behavioral structure, extending beyond variations in the execution of a specified action. Generating appropriate physics-based reactions therefore requires both selecting among these alternatives and maintaining the selected behavior throughout physical execution.

Autoregressive motion diffusion generates successive motion segments conditioned on recent history and external inputs~\cite{tevet2025closd,zhao2025dartcontrol}. In our continuous-only variant, behavioral choice remains implicit in local denoising conditioned on the cue and execution history. We hypothesize that this formulation makes reaction selection and persistence more difficult when heterogeneous behaviors share similar initial states. Recent history describes what has already occurred but may not sufficiently specify how the complete reaction should unfold. DisCo-MoT addresses this ambiguity by sampling a complete discrete reaction plan and retaining its context across prediction windows. The continuous branch then predicts detailed future states under this persistent guidance while adapting to observed execution history.

The ablation results in Table~3 of the main paper support the contribution of this global branch. Removing it reduces MModality from 1.320 to 1.001 for text, from 1.324 to 0.816 for audio, and from 1.244 to 1.101 for video, while worsening FID across all three modalities. These results support explicit whole-reaction modeling as a useful complement to local continuous prediction.

The CGD ablation in Table~\ref{table:discrete_global} further distinguishes the contribution of discrete modeling from that of global context. We replace the discrete global branch with Continuous Global Diffusion (CGD), retaining complete global plans, the same parameter count, and the same local autoregressive diffusion branch. Despite preserving global context, CGD yields lower MModality: 1.040 versus 1.320 for text, 1.086 versus 1.324 for audio, and 1.103 versus 1.244 for video. MixReact also achieves higher Diversity across all three modalities, comparable text/audio FID, and better video FID (0.706 versus 0.768). Thus, the diversity gains extend beyond the availability of global planning and support the advantage of its discrete formulation.

Together, these ablations support our motivation for discrete global modeling: explicitly selecting among distinct behavioral alternatives helps mitigate the tendency toward mode collapse in one-to-many cue--reaction generation, while local diffusion preserves the capacity for fine-grained continuous realization.

\section{Limitations and Future Work}\label{limitations}

MixReact currently generates reactions in physics-based simulation without explicitly modeling the spatial relationships and contact requirements between the humanoid and the entities depicted in the cue. For example, a petting gesture elicited by a dog video does not account for the animal's relative position or enforce appropriate hand--animal contact. Extending reaction generation to physically grounded interactions requires integrating scene geometry, interaction targets, and contact constraints into planning and control. Such an extension would allow the same reaction intent to be realized through motions adapted to different target locations and environmental configurations. Deployment on real humanoids additionally introduces challenges involving perception uncertainty and differences between simulated and real dynamics. Future work could investigate domain randomization and online adaptation to improve robustness to sensing errors and dynamics mismatch.

Although MixReact supports text, audio, and video within a unified framework, each reaction uses a single cue modality because the datasets used in this work lack jointly annotated cues from all three modalities for the same reaction. Future work could collect high-quality datasets pairing synchronized multimodal cues with multiple plausible reactions, enabling models to integrate complementary signals while preserving response diversity. Learning to resolve conflicting cues and update ongoing behavior as new information becomes available would further extend MixReact toward continuous interaction in dynamic environments.

\end{document}